%% file: main_final.tex
 \pdfoutput=1
\documentclass[10pt,twocolumn,letterpaper]{article}

\usepackage{cvpr}   

\usepackage{graphicx}
\usepackage{amsmath}
\usepackage{amssymb}
\usepackage{booktabs}
\usepackage{mathptmx}

\usepackage[utf8]{inputenc} 
\usepackage[T1]{fontenc}    
\usepackage{url}            
\usepackage{multirow}
\usepackage{amsfonts}       
\usepackage{nicefrac}       
\usepackage{microtype}      
\usepackage{adjustbox}
\usepackage{color}
\usepackage[toc,page]{appendix}
\usepackage{wrapfig,lipsum,booktabs}
\usepackage{wasysym}
\usepackage{amsfonts}
\usepackage{xcolor}         

\usepackage[pagebackref,breaklinks,colorlinks]{hyperref}

\usepackage[capitalize]{cleveref}
\crefname{section}{Sec.}{Secs.}
\Crefname{section}{Section}{Sections}
\Crefname{table}{Table}{Tables}
\crefname{table}{Tab.}{Tabs.}

\def\cvprPaperID{5922} 
\def\confName{CVPR}
\def\confYear{2022}

\begin{document}

\title{Enhancing Classifier Conservativeness and Robustness by Polynomiality}

\author{Ziqi Wang$^\bullet$\\
$^\bullet$Delft University of Technology\\
The Netherlands\\
{\tt\small z.wang-8@tudelft.nl}
\and
Marco Loog$^{\circ\bullet}$\\
$^\circ$University of Copenhagen\\
Denmark\\
{\tt\small m.loog@tudelft.nl}
}
\maketitle

\begin{abstract}    
We illustrate the detrimental effect, such as overconfident decisions, that exponential behavior can have in methods like classical LDA and logistic regression. We then show how polynomiality can remedy the situation.  This, among others, leads purposefully to random-level performance in the tails, away from the bulk of the training data. A directly related, simple, yet important technical novelty we subsequently present is \emph{softRmax}: a reasoned alternative to the standard softmax function employed in contemporary (deep) neural networks. It is derived through linking the standard softmax to Gaussian class-conditional models, as employed in LDA, and replacing those by a polynomial alternative. We show that two aspects of softRmax, conservativeness and inherent gradient regularization, lead to robustness against adversarial attacks without gradient obfuscation.

\end{abstract}

\input{introduction}

\input{related_works}

\input{method}
\input{experiments}

\input{conclusion}
\input{acknowledgement}
{\small
\bibliographystyle{ieee_fullname}
\bibliography{egbib}
}
\end{document}

%% file: introduction.tex
\section{Introduction}\label{section: Introduction}
Models
that show some form of exponential behavior are ubiquitous in machine learning: from the Gaussian class conditional distribution in linear discriminant analysis (LDA) \cite{fisher1936use,hastie1996discriminant} to sigmoid activation for logistic regression \cite{kleinbaum2002logistic, menard2002applied}, and the softmax activation function in deep neural networks \cite{krizhevsky2012imagenet, simonyan2014very}. 
Models with such use of exponentiality can, however, have unwanted behavior. We describe and illustrate such behavior, examine its reason, and propose a partial remedy by switching to models that behave polynomially.  Like \cite{hill1975simple, chen1978remark}, we consider the distribution tail and show that samples in the tails receive overconfident posterior predictions \cite{kristiadi2020being}. This renders the model sensitive to outliers and causes overfitting, especially in the case of distribution shift. Moreover, we link overconfident predictions to the lack of robustness against gradient based adversarial attacks.



A model should not be certain about a sample that deviates too much from the training data. Overconfident predictions on samples in the distribution tails should often be avoided, e.g.\ an atypical patient may otherwise be classified to be healthy or diseased with strong confidence.  We want what we call \emph{conservativeness}, which expresses the fact that we are uncertain.  Specifically, we define it to be random guess-level prediction for samples in the tail of the distribution and show that this can be achieved by moving from exponential to polynomial behavior both in LDA and logistic regression.
In addition, for logistic regression and deep learning, studies into the standard softmax activation have shown that it is not necessarily the best choice in many settings \cite{de2016exploration, kanai2018sigsoftmax, titsias2016one}. We propose a polynomial form of softmax posterior estimation that we coin \emph{softRmax}. For this, we exploit the connection between the standard softmax function and LDA \cite{bishop1995neural} and adopt a modified Cauchy distribution as the substitute for the (super)exponential Gaussian term. 

Besides overconfident predictions, the use of exponentiality is also linked to vulnerability to adversarial attacks. Such attacks aim to cause malicious prediction changes by adding an unnoticeable perturbation to the original input. \emph{Robustness} is the ability to maintain performance under adversarial attacks \cite{carlini2019evaluating}.
We demonstrate that a higher robustness of neural networks can be obtained by simply substituting the standard softmax with our softRmax. We show that the robustness can be linked back to the conservativeness of softRmax and  inherent gradient regularization. The first factor, conservativeness, mainly brings robustness against gradient based attacks. 
The second leads to an enlarged margin between samples and the decision boundary, thereby boosting robustness against attacks as well. The effectiveness of various strategies countering adversarial attacks can be attributed to gradient obfuscation \cite{finlay2021scaleable, athalye2018obfuscated}. We show that our inherent gradient regularization does not rely on such obfuscation. 



We sketch the benefits of conservativeness under covariate shift \cite{sugiyama2007covariate,sugiyama2008direct, gretton2009covariate} and show it when a model is under attack. We verify the robustness of our polynomial substitutes empirically on toy and public datasets. 
We further propose a semi-black-box attack, which we call an average-sample attack, to confirm that the robustness of our softRmax indeed comes from the above two factors. 
We also introduce a scale-invariant metric, the magnitude-margin ratio, for comparing the robustness of different models under the same level of attack.

%% file: related_works.tex
\section{Background Material and Related Methods}\label{section: related work}

Adversarial attacks are used for robustness evaluation in our work. They are categorized into white-box and black-box attacks, depending on whether the network is available or not \cite{papernot2016towards}. Black-box attacks do not need the network architecture and usually involve the training of a substitute network that mimics the decision boundary of the target network \cite{papernot2017practical}. A gradient-based adversarial attack is a typical white box attack \cite{goodfellow2015explaining, kurakin2016physical}. It aims to find the perturbation direction that can lead to the fastest change in the prediction.

FGSM \cite{goodfellow2015explaining} is a simple yet effective approach where a small perturbation $\mathbf{\eta}$ is added to the input $\mathbf{x}$ to increase the overall loss. The perturbation $\mathbf{\eta}$ is $\epsilon$ multiplied by the sign of the loss gradient $\nabla_xJ(\mathbf{w}, \mathbf{x}, y)$.  The perturbed input becomes:
\begin{equation}\label{eq: FGSM}
    \mathbf{x}' = \mathbf{x} + \epsilon \,  \text{sign}(\nabla_xJ(\mathbf{w}, \mathbf{x}, y)).
\end{equation}
Similarly, a gradient-based target attack \cite{papernot2016limitations} aims to perturb the sample to a target class $y_t$ by decreasing the loss that corresponds to the target class:
\begin{equation}
    \mathbf{x}' = \mathbf{x} - \epsilon \, \text{sign}(\nabla_xJ(\mathbf{w}, \mathbf{x}, y_t)).
\end{equation}
BIM \cite{kurakin2016physical} performs the attack iteratively in $T$ steps. With the same attacking scale $\epsilon$, BIM applies the attack at the scale of $\alpha = \epsilon/T$ in each step to form an attacked input $\mathbf{x}'_t$ at step $t$: 
\begin{equation}\label{eq: BIM}
    \mathbf{x}'_{t+1} = \mathbf{x}'_t + \alpha \, \text{sign}(\nabla_xJ(\mathbf{w}, \mathbf{x}, y)).
\end{equation}


We need the notion of a prediction margin $M_z$ \cite{tsuzuku2018lipschitz} to measure the robustness to adversaries, which has an indirect link to the classical (geometrical) margin in the input space \cite{ScholkopfS02}. Our work uses it to evaluate the margin and the robustness of our method. For this, we consider a mapping from the input $\mathbf{x}$ to the latent or representation space: $\mathbf{z} = f(\mathbf{x}, \mathbf{w})$, with $\mathbf{z}\in \mathbb{R}^{k}$ and $\mathbf{z}_i$ the output of the final layer corresponding to class $i\in \{0,1,\ldots,k\}$. Assuming a sample $\mathbf{x}$ is correctly classified to its class $y$, $\mathbf{z}_y$ takes on the maximum value in $\mathbf{z}$. The prediction margin is defined as the distance between $\mathbf{z}_y$ and the second largest value in $\mathbf{z}$:
\begin{equation}\label{eq: prediction margin}
    M_z := z_y - \max_{i\neq y}\{z_i\}.
\end{equation}

Adversarial defenses for deep learning have been achieved by adversarial training \cite{kurakin2016adversarial}, distillation \cite{ba2014deep,papernot2016distillation}, constructing a maximum margin in the latent space \cite{pang2018max, pang2019rethinking, wan2018rethinking, hess2020softmax} and gradient regularization \cite{drucker1992improving,ross2018improving,hess2020softmax,tsuzuku2018lipschitz,nayebi2017biologically,goroshin2013saturating}. Explanations for gradient regularization approaches are heuristic and their successes often hinge on gradient obfuscation \cite{finlay2021scaleable}. The latter refers to an unnecessarily rough loss landscape that hinders gradient-based adversarial attacks, which get readily stuck in the local minima of the roughened loss. It should be noted, however, that this approach does not solve the problem of adversarial attacks inherently \cite{athalye2018obfuscated}. Increasing the iteration number in BIM attacks \cite{kurakin2016physical} and using black-box attacks are standard to detect gradient obfuscation. We use both in our work to show that our approach does not rely on gradient obfuscation.

Covariate shift is a specific problem within domain adaptation. Domain adaptation refers to the scenario where the training data and the test data are not i.i.d. \cite{daume2009frustratingly, kouw2019review}. The training and test data are referred to as the source domain and the target domain, respectively. One standard solution of this problem is to approximate the target domain by assigning the source samples weights determined by the source and target distribution. In the original work \cite{sugiyama2007covariate}, these are estimated using Gaussian distributions.  With this approach, adding very few samples in the tail of the source distribution can lead to considerable overfitting to the added outliers. We illustrate that, if a polynomial $t$-distribution instead of Gaussian distribution is adopted in the procedure of density estimation, the influence of outliers is limited.

%% file: method.tex
\section{Exponentiality vs Polynomiality}\label{section: Method}
We first demonstrate the presence of overconfident prediction in the distribution tail and sensitivity to adversarial attacks with classical LDA, logistic regression, and deep learning. We then replace the exponential terms in each scenario by polynomial ones and show that this substitution is a simple yet effective approach to deliver conservativeness and improved robustness. Notably, for the latter, no adversarial training or extra regularization is required.
\subsection{Conservativeness}

Conservativeness is defined as estimating the posterior class probabilities $p(y_i|\mathbf{x})$ at random-guess level for $\mathbf{x}$ in the tail, away from the bulk of the data.
To study such tail behavior, we basically study $\mathbf{x}$ for which the norm grows indefinitely, i.e., $\|\mathbf{x}\|\to\infty$. Assuming $k$ classes and ignoring class priors, conservativeness comes down to the requirement that we can informally state as:
\begin{equation}\label{eq: conservative definition}
    \lim_{\|\mathbf{x}\|\to \infty}p(y_i|\mathbf{x}) \approx \frac{1}{k}.
\end{equation}

\subsubsection{LDA}\label{section: Bayes classifier}

\begin{figure}
  \centering
  \begin{subfigure}{0.45\linewidth}
    \includegraphics[width = 1 \textwidth]{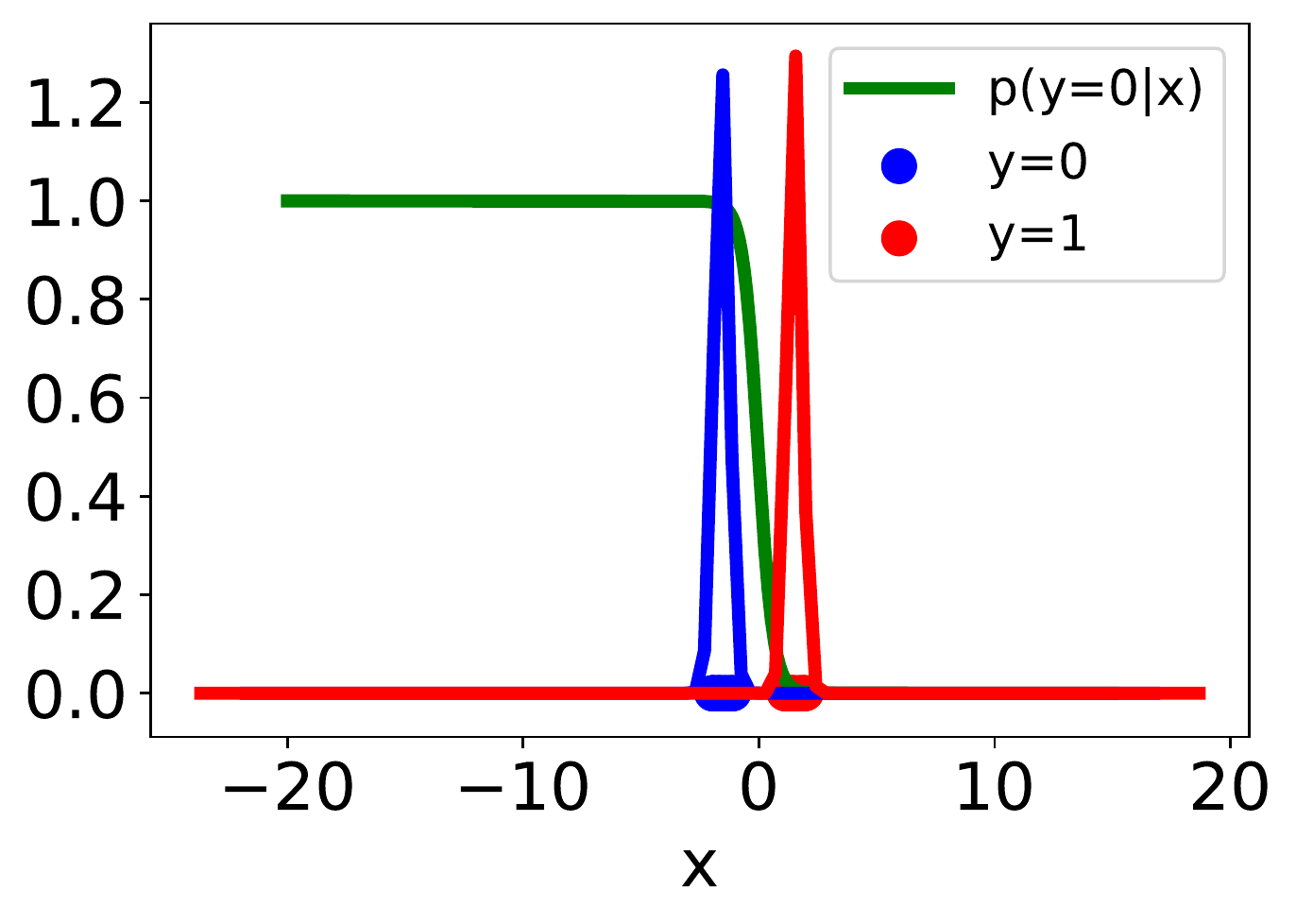}
    \caption{Gaussian distribution.}
     \label{subfig:NB_Gaussian}
  \end{subfigure}
  \begin{subfigure}{0.45\linewidth}
    \includegraphics[width = 1 \textwidth]{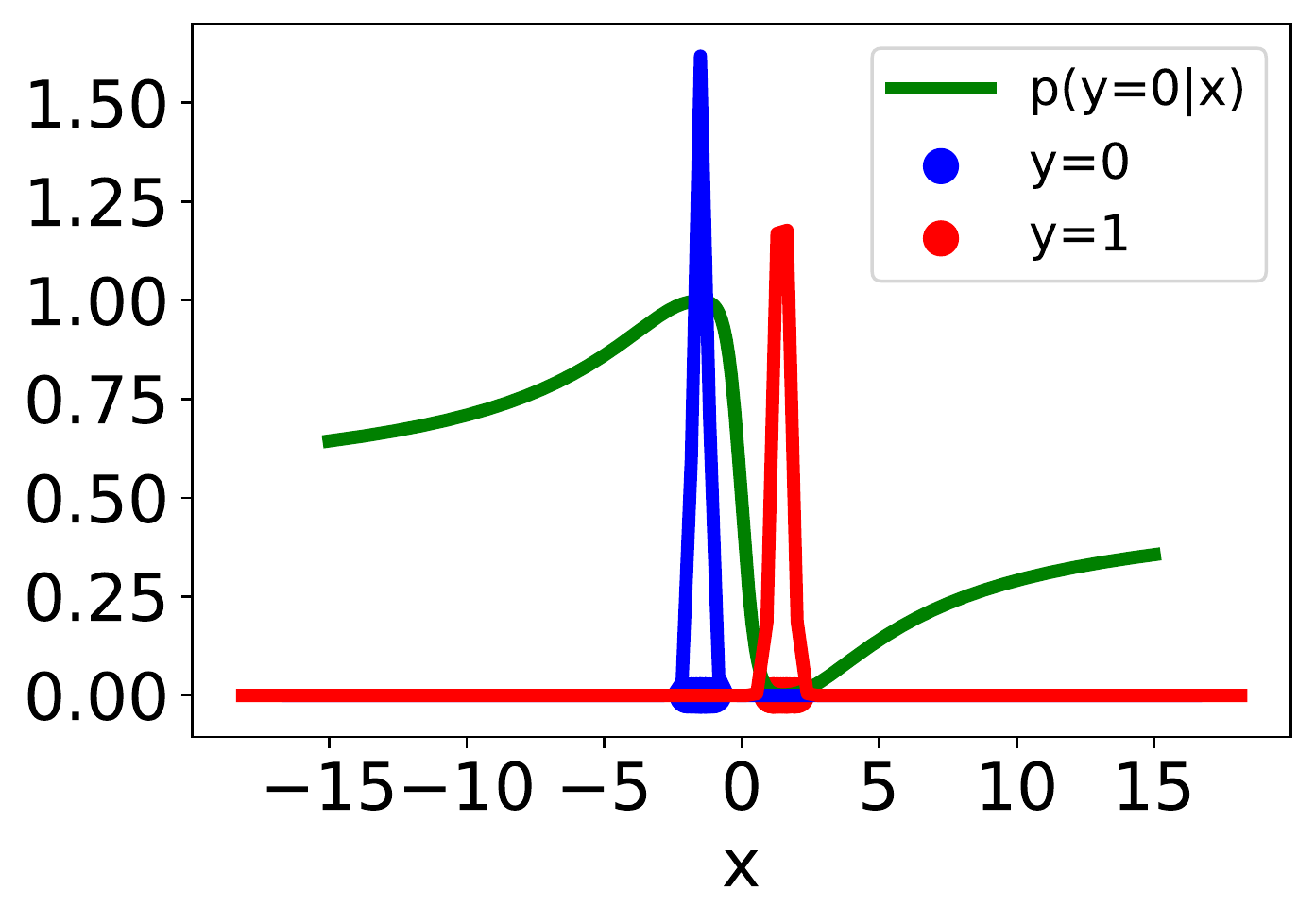}
    \caption{Student's $t$-distribution.}
    \label{subfig: NB_t}
  \end{subfigure}
  \begin{subfigure}{0.45\linewidth}
    \includegraphics[width = 1 \textwidth]{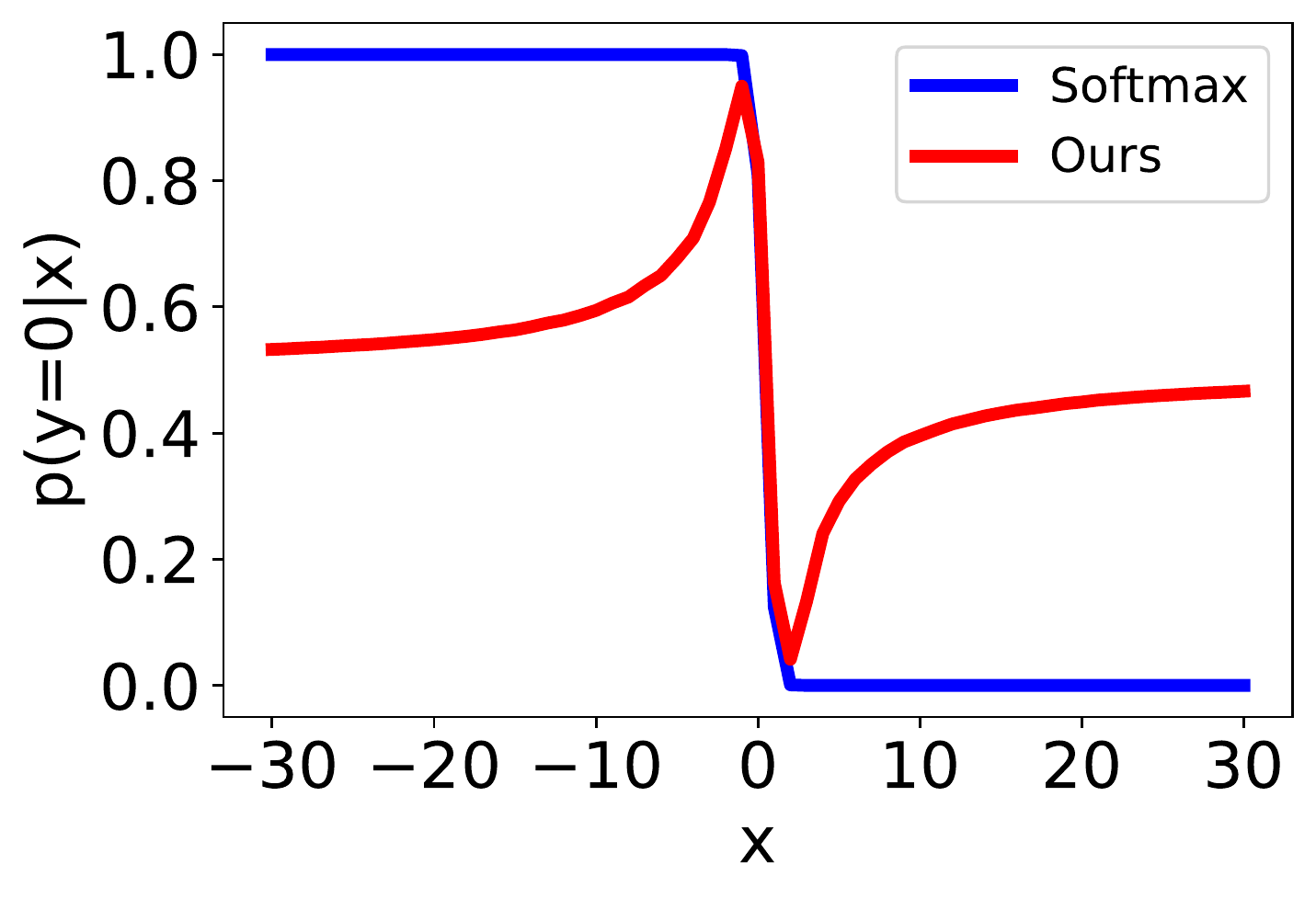}
    \caption{Logistic regression.}
    \label{subfig: regression}
  \end{subfigure}
  \caption{LDA and logistic regression with exponential and polynomial assumptions. Posteriors $p(y_0|x)$ are compared. Subfigure \ref{subfig:NB_Gaussian} and \ref{subfig: NB_t} show the predicted posterior by LDA with class conditional Gaussian or $t$-distributions. Subfigure \ref{subfig: regression} compares the posterior of softmax and softRmax. LDA with Gaussian assumption and softmax with exponential functions show overconfident predictions in the distribution tails. Conservative prediction is achieved by substituting polynomial for exponential behavior.}
 \label{fig: conservativeness}
\end{figure}

We consider $k$-class classification using LDA. 
We elaborate upon the link between the overconfident prediction and exponentiality.  Following Bayes' rule, the posterior of class $y_i$, under equal priors, is
\begin{equation}\label{eq:bayes}
    p(y_i|\mathbf{x}) = \frac{p(y_i)p(\mathbf{x}|y_i)}{\sum_k p(y_k)p(\mathbf{x}|y_k)} = \frac{p(\mathbf{x}|y_i)}{\sum_k p(\mathbf{x}|y_k)}.
\end{equation}
Consider $\mathbf{x}$ to be 1D for simplicity.
The class conditional distribution $p(x|y)$ is estimated by fitting a Gaussian  $N(x|\mu_k,\sigma^2)$ with $\mu_k$ and $\sigma^2$ being the mean and variance of class $k$. 
When $x$ goes to $\pm$infinity, the posterior saturates to one-hot encoding due to the (faster than) exponential rate of decrease of the Gaussian distribution.  Specifically, we have
\begin{align}
& p(y_i|x) = \label{eq:other} \\ 
& \left( 1 + \sum_{k\neq i} \exp \left( -\frac{1}{2 \sigma^2}(2x(\mu_i-\mu_k)-\mu_i^2+\mu_k^2) \right) \right)^{-1}\nonumber
\end{align}
from which we see that $\lim_{x\to\pm\infty} p(y_i|x)=0$, unless $y_i$ is the mean closest to $x=\pm\infty$, in which case the posterior will be $1$.
This is also illustrated in Figure \ref{subfig:NB_Gaussian}.

\paragraph{Polynomial substitute.}
We propose to substitute the Gaussian distribution with the (noncentral) Student's $t$-distribution in the density estimation. Other distributions that fall of polynomially can be considered as long as the power of the leading terms are the same for all $k$ class conditional distributions. In this way, conservative posteriors with a behavior as in Equation \eqref{eq: conservative definition} are obtained.

The reason for this is that the limit of $x$ going to $\pm$infinity for Equation \eqref{eq:bayes} behaves rather different when the numerator and denominator contain polynomial instead of exponential terms.  For the former, convergence is controlled by the polynomial decay rate of the posteriors $p(x|y_k)$.  When equal, the limit posterior, assuming all priors equal, is $\tfrac{1}{k}$.

\paragraph{Example.}
We consider a binary classification task in 1D data. We assume a uniform distribution in the range $[-2, -1]$ for class $y_0$ and $[1, 2]$ for class $y_1$. With Gaussian distributions for the class conditional distributions $p(x|y_0)$ and $p(x|y_1)$---fitted using maximum likelihood, we get the change of posterior $p(x|y_0)$ w.r.t. input $\mathbf{x}$ as in Figure \ref{subfig:NB_Gaussian}. 
When $x\to-\infty$, $p(y_0|x)=1$ and when $x\to \infty$ $p(y_1|x)=1$. Substituting the $t$-distribution for the Gaussian, as shown in Figure \ref{subfig: NB_t}, for samples that are in the bulk of the class conditional distribution, we still obtain a posterior $p(y_0|x)$ close to 1. But for samples in the tail, we find more conservative prediction where $p(y_0|x)$ and $p(y_1|x)$ are approximately $\tfrac{1}{2}$. 

\subsubsection{Logistic Regression and Softmax}\label{section: method_logistic regression}
The softmax in neural network, as employed in the last layer to come to posterior estimates, works in the same way as multi-class logistic regression for classification tasks. Here, we consider a basic linear transformation $f(\mathbf{x}, \mathbf{w}) = \mathbf{w}^T\mathbf{x} + \mathbf{b} = \mathbf{z}$, though our analysis can be readily generalized to nonlinear neural networks.

With the standard softmax activation $\varsigma^S$, the embedding $
\mathbf{z}$ is mapped to a vector of posteriors with $p(y_i|\mathbf{x}) = \varsigma_i^S(\mathbf{z}) =e^{\mathbf{z}_i}/\sum_ke^{\mathbf{z}_k}$.  Equivalent to Equation \eqref{eq:other}, we have
\begin{equation}\label{eq: regression infinity} 
\begin{split}
    &\varsigma_i^S(\mathbf{z})=\frac{\exp(\mathbf{w}_i^T\mathbf{x} + \mathbf{b}_i)}{\sum_k \exp(\mathbf{w}_k^T\mathbf{x} + \mathbf{b}_k)}= \\
    & \left( 1 + \sum_{k\neq i}\exp \left( (\mathbf{w}_k-\mathbf{w}_i)^T\mathbf{x} + (\mathbf{b}_k - \mathbf{b}_i) \right) \right)^{-1}. 
\end{split}
\end{equation}
When $\|\mathbf{x}\|\to \infty$, if $(\mathbf{w}_k - \mathbf{w}_i)^T\mathbf{x}$ is negative for all $k, k \neq i$, then the posterior will be 1, otherwise it is 0.

We make the connection of softmax with LDA here. Let us position $k$ normal distributions with identity covariance, $N(\cdot|\mathbf{m},\mathbf{I})$, in $Z$. 
Their means are the $k$ standard basis vector $\mathbf{e}_k$. 
Based on these distributions---every single one of them representing one of the $k$ classes, we can map every $\mathbf{z} \in Z$ to a vector of posteriors $\varsigma^G(\mathbf{z})$, simply by setting
\begin{equation}\label{eq: fit_Gaussian_softmax_conditional}
    \varsigma_i^G(\mathbf{z}) := \frac{N(\mathbf{z}|\mathbf{e}_i,\mathbf{I})}{\sum_k N(\mathbf{z}|\mathbf{e}_k,\mathbf{I})}.
\end{equation}
This, in turn, can be directly related to the softmax $\varsigma^S$.  First, we realize that, for $\mathbf{z}$ fixed,
\begin{equation}
\begin{split}
&N(\mathbf{z}|\mathbf{e}_i,\mathbf{I}) \propto \exp(-\tfrac{1}{2} \|\mathbf{z}-\mathbf{e}_i\|^2)\label{eq: approximate_softmax_Gaussian_conditional}\\
\propto & \exp\left(-\frac{1}{2}\sum_{k} \mathbf{z}_k^2\right) \exp(\mathbf{z}_i) \exp(-\tfrac{1}{2}) \propto e^{\mathbf{z}_i}.
\end{split}
\end{equation}
From this, we immediately see that
\begin{equation}\label{eq: approximate_softmax}
    \varsigma_i^G(\mathbf{z}) = \frac{N(\mathbf{z}|\mathbf{e}_i,\mathbf{I})}{\sum_k N(\mathbf{z}|\mathbf{e}_k,\mathbf{I})} = \frac{e^{\mathbf{z}_i}}{\sum_k e^{\mathbf{z}_k}} = \varsigma_i^S(\mathbf{z}).
\end{equation}

\paragraph{Polynomial substitute.}
Inspired by the standard Cauchy distribution $p_C(x) = \frac{1}{\pi(1+x^2)}$---a specific $t$-distribution, we use a polynomial term with the power of $-2$ to substitute the Gaussian class conditional distribution $N(\mathbf{z}|\mathbf{e}_i,\mathbf{I})$ in Equation \eqref{eq: approximate_softmax}, which gives our \emph{softRmax} activation function $\varsigma^C$:
\begin{equation}\label{eq: our_softmax}
    \varsigma_i^C(\mathbf{z}) := \frac{\frac{1}{\|\mathbf{z}-\mathbf{e}_i\|^2}}{\sum_k \frac{1}{\|\mathbf{z}-\mathbf{e}_k\|^2}}.
\end{equation}
By adopting the polynomial function, the posterior becomes conservative, because
\begin{equation}\label{eq: regression posterior polynomial}
\begin{aligned}
p(y_i|\mathbf{x}) =  \varsigma_i^C(\mathbf{z})&= \frac{\|\mathbf{w}^T\mathbf{x} + \mathbf{b}-\mathbf{e}_i\|^{-2}}{\sum_k \|\mathbf{w}^T\mathbf{x} + \mathbf{b}-\mathbf{e}_k\|^{-2}}\\
&=\frac{1}{1 + \sum_{k\neq i} \left\|\frac{\mathbf{w}^T\mathbf{x} + \mathbf{b} - \mathbf{e}_i}{\mathbf{w}^T\mathbf{x }+\mathbf{b} -\mathbf{e}_k}\right\|^{2}}
\end{aligned}
\end{equation}
and the terms $\left\|\frac{\mathbf{w}^T\mathbf{x} + \mathbf{b} - \mathbf{e}_i}{\mathbf{w}^T\mathbf{x }+\mathbf{b} -\mathbf{e}_k}\right\|^{2}$ converge to 1 when $\|\mathbf{x}\|\to \infty$.
\paragraph{Example.}
We consider logistic regression for binary classification in 1D. Similar to the previous example, we assume a uniform distributions in the ranges $[-1,0]$ and $[1,2]$ for the two classes $y_0$ and $y_1$. The sigmoid/softmax function is substituted with the softRmax activation function from Equation \eqref{eq: our_softmax} to construct a conservative regressor. In Figure \ref{subfig: regression}, we see that the posterior $p(y_0|x)$ goes to $\tfrac{1}{2}$ on both ends.

\subsection{Robustness}\label{section: Robustness}

Next to softRmax being conservative, simply substituting the standard softmax with it in any probabilistic deep net also brings more adversarial robustness. We show that this comes from conservativeness in the tail and an inherent weight regularization that leads to an enlarged margin between samples and the decision boundary.
 \begin{figure*}[hbt]
  \centering
  \begin{subfigure}{0.14\linewidth}
    \includegraphics[width = 1 \textwidth]{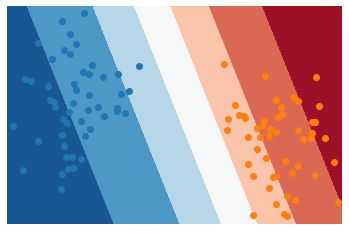}
    \caption{Softmax 1}
     \label{subfig: linear_softmax_epoch1}
  \end{subfigure}
  \hfill
  \begin{subfigure}{0.14\linewidth}
    \includegraphics[width = 1 \textwidth]{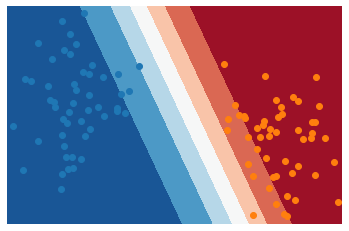}
    \caption{Softmax 5}
     \label{subfig: linear_softmax_epoch5}
  \end{subfigure}
  \hfill
  \begin{subfigure}{0.14\linewidth}
    \includegraphics[width = 1 \textwidth]{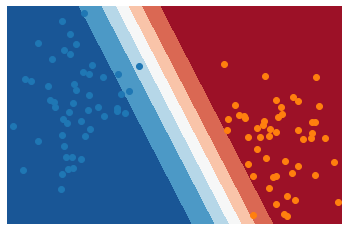}
    \caption{Softmax 10}
     \label{subfig: linear_softmax_epoch10}
  \end{subfigure}
  \hfill
  \begin{subfigure}{0.14\linewidth}
    \includegraphics[width = 1 \textwidth]{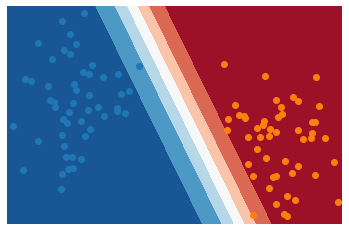}
    \caption{Softmax 15}
     \label{subfig: linear_softmax_epoch15}
  \end{subfigure}
  \hfill
  \begin{subfigure}{0.14\linewidth}
    \includegraphics[width = 1 \textwidth]{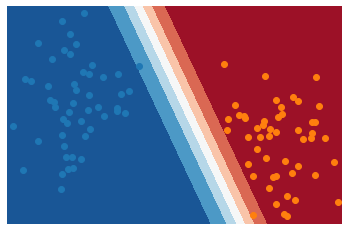}
    \caption{Softmax 30}
     \label{subfig: linear_softmax_epoch30}
  \end{subfigure}
  \hfill
  \begin{subfigure}{0.14\linewidth}
    \includegraphics[width = 1 \textwidth]{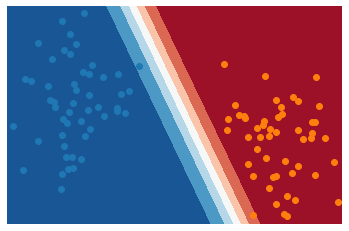}
    \caption{Softmax 50}
     \label{subfig: linear_softmax_epoch50}
  \end{subfigure}
  \hfill
  \begin{subfigure}{0.14\linewidth}
    \includegraphics[width = 1 \textwidth]{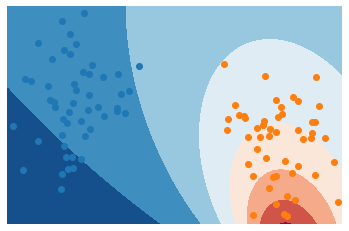}
    \caption{Ours 1}
     \label{subfig: linear_marco_epoch1}
  \end{subfigure}
  \hfill
  \begin{subfigure}{0.14\linewidth}
    \includegraphics[width = 1 \textwidth]{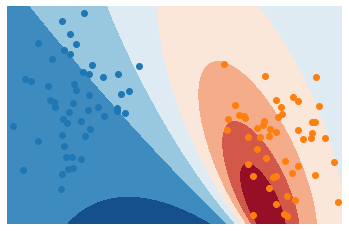}
    \caption{Ours 5}
     \label{subfig: linear_marco_epoch5}
  \end{subfigure}
  \hfill
  \begin{subfigure}{0.14\linewidth}
    \includegraphics[width = 1 \textwidth]{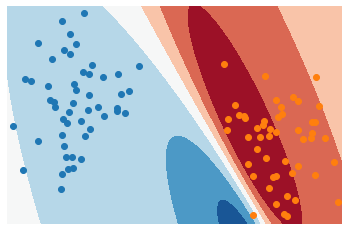}
    \caption{Ours 10}
     \label{subfig: linear_marco_epoch10}
  \end{subfigure}
  \hfill
  \begin{subfigure}{0.14\linewidth}
    \includegraphics[width = 1 \textwidth]{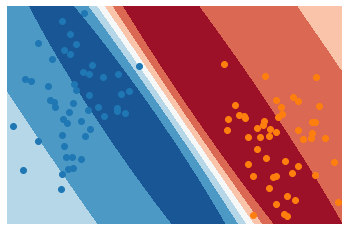}
    \caption{Ours 15}
     \label{subfig: linear_marco_epoch15}
  \end{subfigure}
  \hfill
  \begin{subfigure}{0.14\linewidth}
    \includegraphics[width = 1 \textwidth]{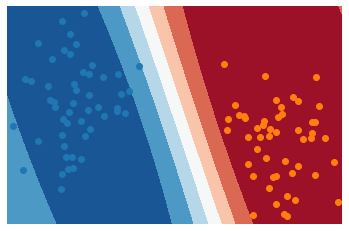}
    \caption{Ours 30}
     \label{subfig: linear_marco_epoch30}
  \end{subfigure}
  \hfill
  \begin{subfigure}{0.14\linewidth}
    \includegraphics[width = 1 \textwidth]{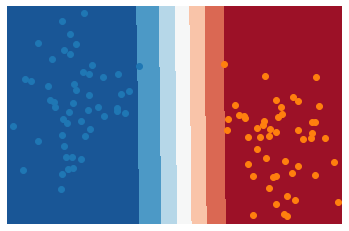}
    \caption{Ours 50}
     \label{subfig: linear_marco_epoch50}
  \end{subfigure}
  \begin{subfigure}{0.14\linewidth}
    \includegraphics[width = 1 \textwidth]{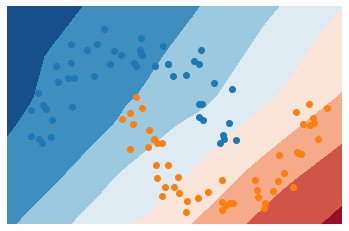}
    \caption{Softmax 1}
     \label{subfig: softmax_epoch1}
  \end{subfigure}
  \hfill
  \begin{subfigure}{0.14\linewidth}
    \includegraphics[width = 1 \textwidth]{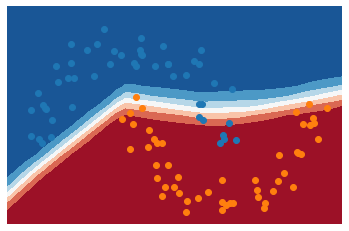}
    \caption{Softmax 10}
     \label{subfig: softmax_epoch10}
  \end{subfigure}
  \hfill
  \begin{subfigure}{0.14\linewidth}
    \includegraphics[width = 1 \textwidth]{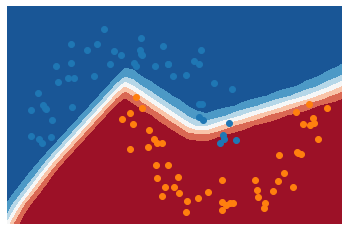}
    \caption{Softmax 15}
     \label{subfig: softmax_epoch15}
  \end{subfigure}
  \hfill
  \begin{subfigure}{0.14\linewidth}
    \includegraphics[width = 1 \textwidth]{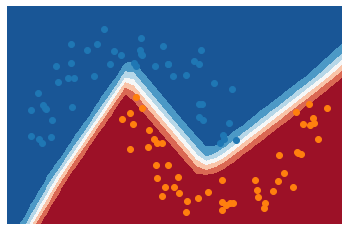}
    \caption{Softmax 20}
     \label{subfig: softmax_epoch20}
  \end{subfigure}
  \hfill
  \begin{subfigure}{0.14\linewidth}
    \includegraphics[width = 1 \textwidth]{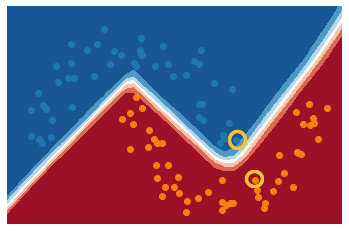}
    \caption{Softmax 30}
     \label{subfig: softmax_epoch30}
  \end{subfigure}
  \hfill
  \begin{subfigure}{0.14\linewidth}
    \includegraphics[width = 1 \textwidth]{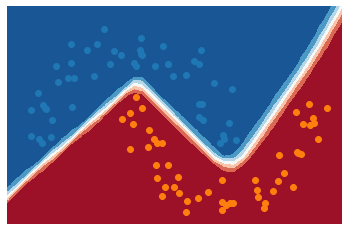}
    \caption{Softmax 50}
     \label{subfig: softmax_epoch50}
  \end{subfigure}
  
  \begin{subfigure}{0.14\linewidth}
    \includegraphics[width = 1 \textwidth]{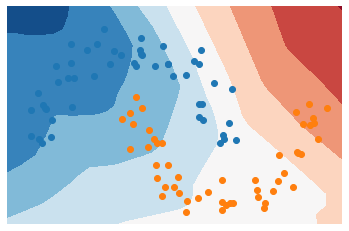}
    \caption{Ours 1}
     \label{subfig: marco_epoch1}
  \end{subfigure}
  \hfill
  \begin{subfigure}{0.14\linewidth}
    \includegraphics[width = 1 \textwidth]{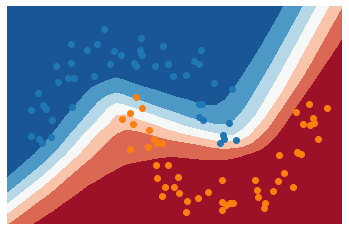}
    \caption{Ours 10}
     \label{subfig: marco_epoch10}
  \end{subfigure}
  \hfill
  \begin{subfigure}{0.14\linewidth}
    \includegraphics[width = 1 \textwidth]{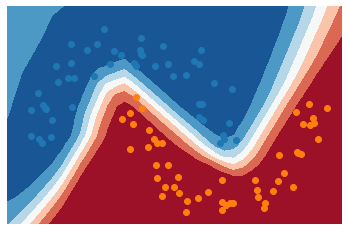}
    \caption{Ours 15}
     \label{subfig: marco_epoch15}
  \end{subfigure}
  \hfill
  \begin{subfigure}{0.14\linewidth}
    \includegraphics[width = 1 \textwidth]{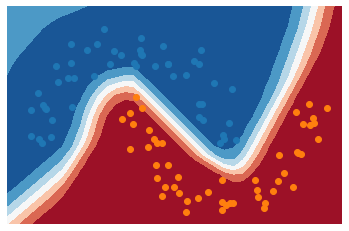}
    \caption{Ours 20}
     \label{subfig: marco_epoch20}
  \end{subfigure}
  \hfill
  \begin{subfigure}{0.14\linewidth}
    \includegraphics[width = 1 \textwidth]{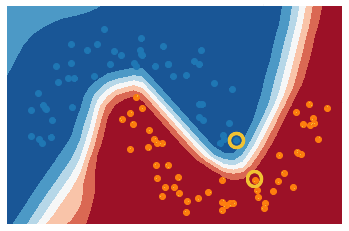}
    \caption{Ours 30}
     \label{subfig: marco_epoch30}
  \end{subfigure}
  \hfill
  \begin{subfigure}{0.14\linewidth}
    \includegraphics[width = 1 \textwidth]{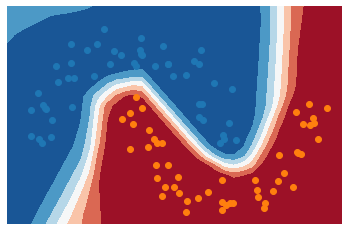}
    \caption{Ours 50}
     \label{subfig: marco_epoch50}
  \end{subfigure}
  \caption{Margin change for linearly separable dataset and the moon dataset with softmax and our softRmax. Different colored points represent the two classes. The color bands show the posterior develops in the input space. The number in the title of each subfigure is the training epoch. With the standard softmax, the model makes the posterior change around the decision boundary sharp to minimize the loss. Due to the regularization of weights $w$ with softRmax, it is harder to minimize the loss by increasing the posterior fast at the decision boundary, which enables the model to find a larger margin.}
\label{fig: linear_softmax-epochs}
\end{figure*}

\subsubsection{Robustness from Conservativeness}\label{section: robustness from the tail}

Most gradient-based adversarial attacks try to maximize the overall loss \cite{goodfellow2015explaining} or minimize the loss of a target class \cite{papernot2016limitations}. For a properly converged network that employs the standard softmax, attacking a correctly classified sample pushes it away from the tail, as the overall loss would not increase moving towards it (and the target class loss would not decrease). 
This is because the posterior of the correct class does not decrease towards the direction of the tail (see Figure \ref{subfig: regression}).  
The loss landscape using softRmax is different due to the conservativeness in the tail, as also illustrated in Figure \ref{subfig: regression}.  For samples that are already positioned in the direction of the tail, an attack would actually push them even further into the tail.  This increases the overall loss or decrease the target class loss. A perturbation towards the tail does, however, not change the accuracy so the attack fails.
This leads a neural network using our softRmax to be more robust to gradient-based attacks. Note that this defense is different from gradient obfuscation because our loss landscape is not unnecessarily rough but simply has a different structure. This will be further elaborated in the corresponding experiment in Subsection \ref{section: exp_gradient_obfuscation}.

\subsubsection{Robustness from Enlarged Margin}
The other factor that contributes to the robustness of softRmax, is the enlarged margin. To illustrate, we again consider a simple linear mapping of the input: $\mathbf{z} = \mathbf{w}^T\mathbf{x} + \mathbf{b}$. The theory can be generalized to nonlinear mappings by substituting the weight $\mathbf{w}$ with the gradient $\nabla_{\mathbf{x}}\mathbf{z}$ in the following derivation. With the output of the activation function being the general $\varsigma$, the posterior gradient equals:
\begin{equation}\label{eq: chain}
    \frac{\partial \varsigma(\mathbf{z})}{\partial \mathbf{x}} = \frac{\partial \varsigma(\mathbf{z})}{\partial \mathbf{z}}\frac{\partial \mathbf{z}}{\partial \mathbf{x}} = \nabla_{\mathbf{z}}\varsigma \mathbf{w}.
\end{equation}

The network weights are optimized by minimizing a posterior based loss function, which means the posterior of the labeled class should be maximized. For a separable dataset, there are many possible decision boundaries that can be learned by the network. When a decision boundary is biased by some samples close to the decision boundary (like in Figures \ref{subfig: linear_softmax_epoch10} and \ref{subfig: softmax_epoch20}), the network generally has two options to further decrease the loss. It can either move the decision boundary to enlarge the classifier margin, or make the transient of posterior steeper at the decision boundary so posteriors of correctly classified samples saturate to 1 quicker. Both of the two approaches decrease the loss.

We observe that with softmax being the activation, the network tends to increase the posterior by making the posterior transient steep (as shown in Figures \ref{subfig: linear_softmax_epoch50} and \ref{subfig: softmax_epoch30}). We believe that this is because the magnitude $m$ of $\mathbf{w}$ is not regularized, so the network can simply increase $m$ during the optimization process to increase the posterior gradient in Equation \eqref{eq: chain}.  This leads to the fast transient of the posterior around the decision boundary. A problem in the optimization is that the posteriors of samples will quickly saturate to 1 and do not contribute to gradient updates anymore. If a decision boundary is biased like in Figure \ref{subfig: linear_softmax_epoch10}, it hardly changes in subsequent epochs. Such decision boundary correctly separates all data, but is more vulnerable to adversarial attacks because the classifier margin is not maximized.

Different from softmax, softRmax optimizes the loss in the other way: by enlarging the margin. It maps $\mathbf{x}$ to $\mathbf{z}$ around the $k$th row of the identity matrix $\mathbf{e}_k$ for class $k$, so $\|\mathbf{z}\| \approx 1$. Correspondingly, the magnitude of weight $\mathbf{w}$ is inherently regularized by $\|\mathbf{w}^T\mathbf{x}+\mathbf{b}\| \approx 1$.
This avoids increasing the weights to values that can lead to a sharp decision boundary and leaves just one of the two above-mentioned optimization options to decrease the loss, i.e., enlarging the margin (see Figs. \ref{subfig: linear_marco_epoch50} and \ref{subfig: marco_epoch30}), resulting in increased robustness. 

%% file: experiments.tex
\section{Experiments}\label{section: Experiments}
We present experimental results on conservativeness and robustness when using standard exponential terms and polynomial substitutes respectively. 
First, we use covariate shift adaptation by importance weighting with outliers as an example to demonstrate that the conservativeness brought by polynomiality is necessary in an LDA-like setting. A next experiment shows that, even under attack, softRmax gives conservative posteriors. 
We also perform standard adversarial attacks on public datasets 
to compare the robustness of softmax and softRmax. To better understand the behavior of softRmax, we introduce a new, so-called, average-sample attack and the magnitude-margin ratio.
\subsection{Conservativeness}\label{section: covariate shift}
\subsubsection{Covariate Shift}
\begin{figure*}[h]
  \centering
  \begin{subfigure}{0.19\linewidth}
    \includegraphics[width = 1 \textwidth]{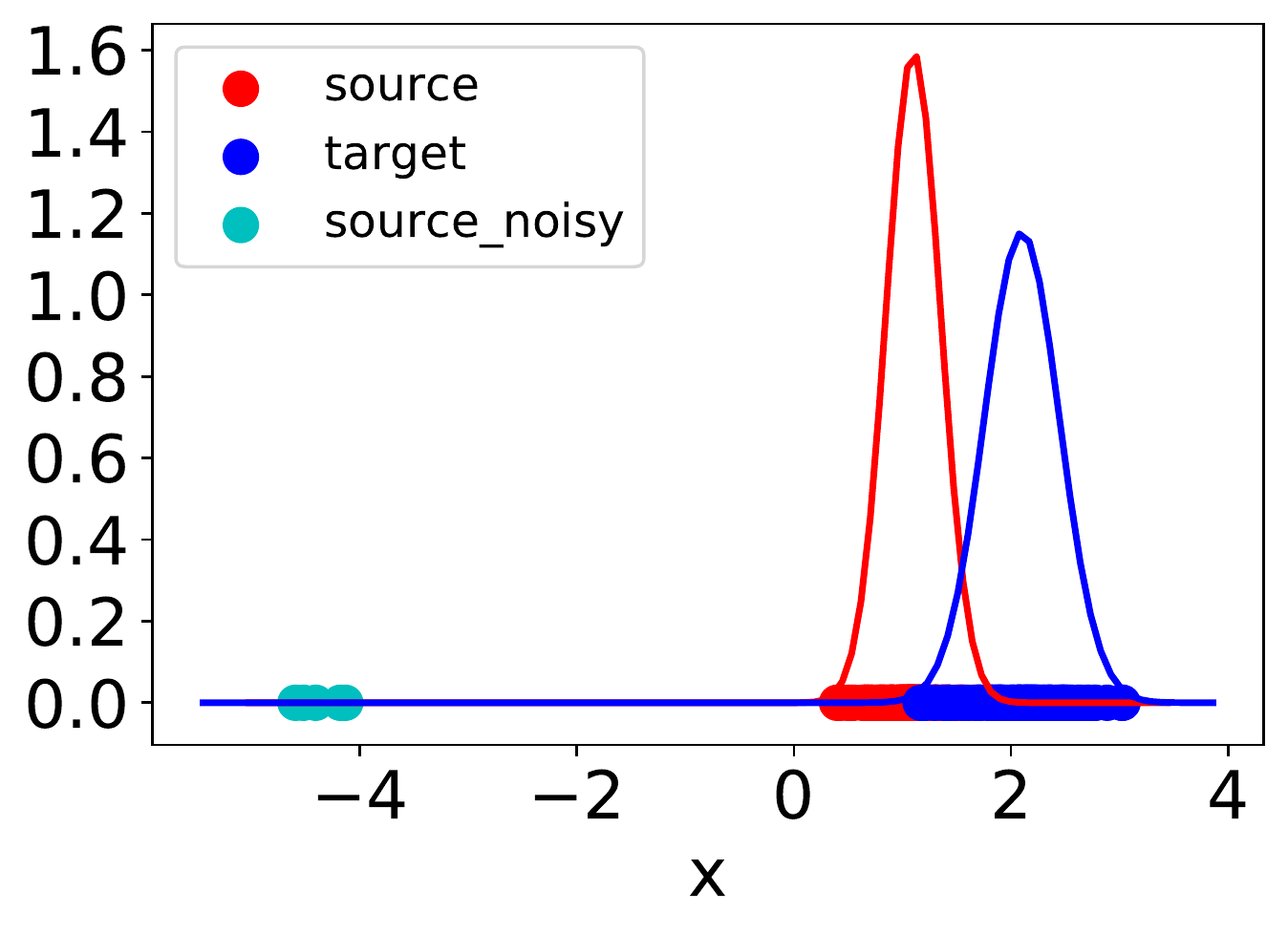}
    \caption{Marginal distribution}
     \label{marginal}
  \end{subfigure}
  \begin{subfigure}{0.19\linewidth}
    \includegraphics[width = 1 \textwidth]{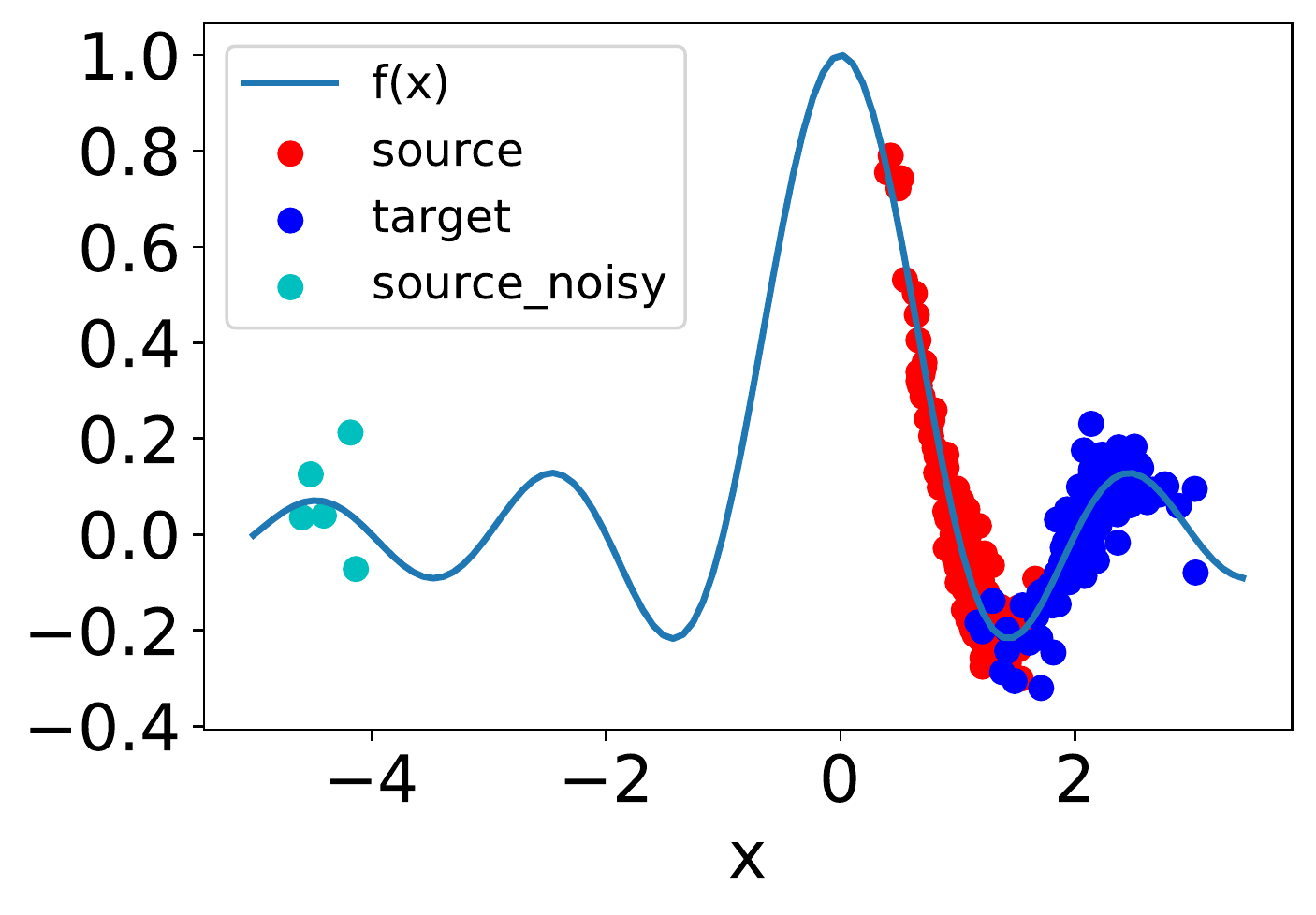}
    \caption{Target function}
    \label{target_function}
  \end{subfigure}
  \begin{subfigure}{0.19\linewidth}
    \includegraphics[width = 1 \textwidth]{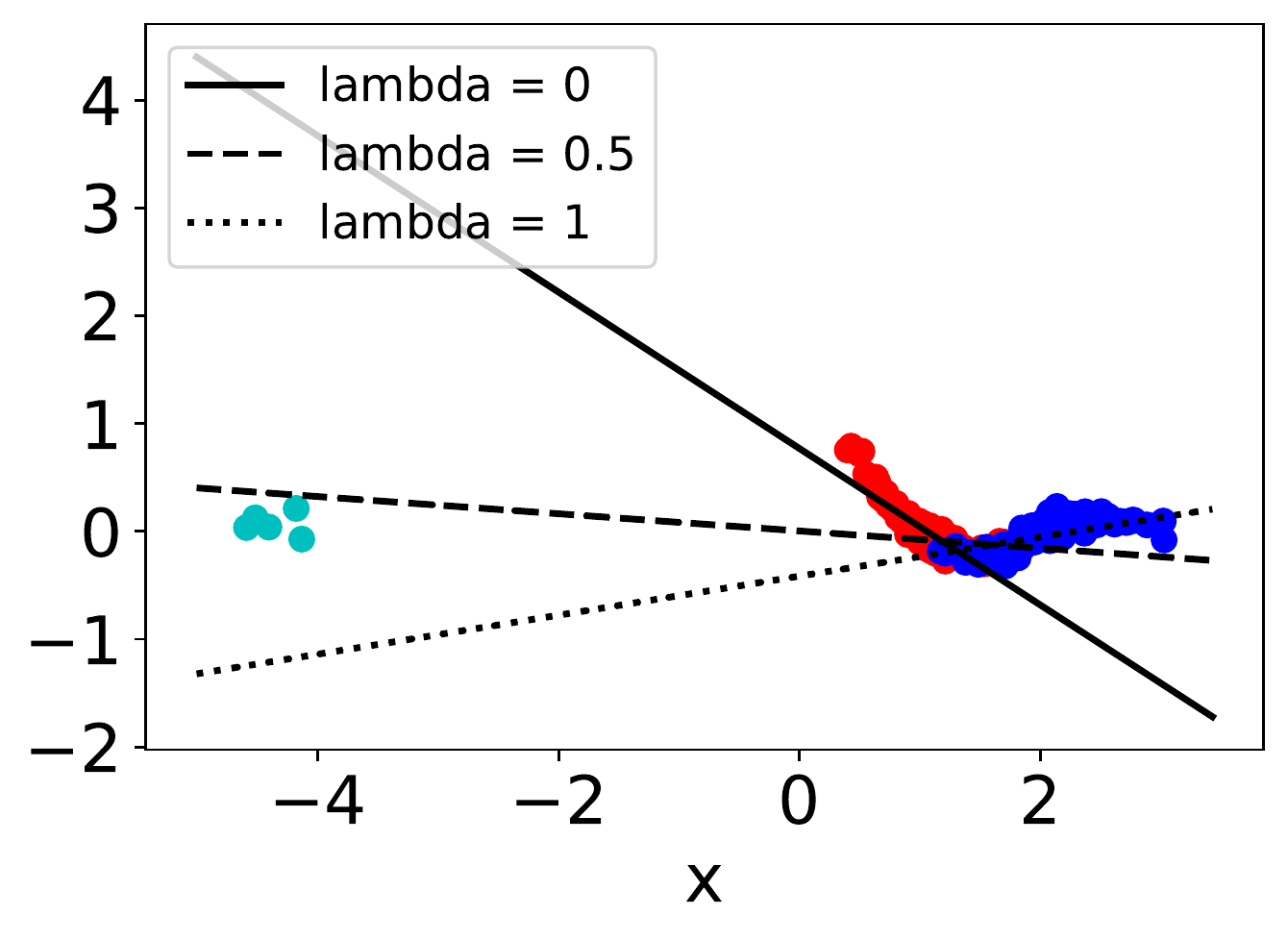}
    \caption{Wihtout outliers}
    \label{no_noise}
  \end{subfigure}
  \begin{subfigure}{0.19\linewidth}
    \includegraphics[width = 1 \textwidth]{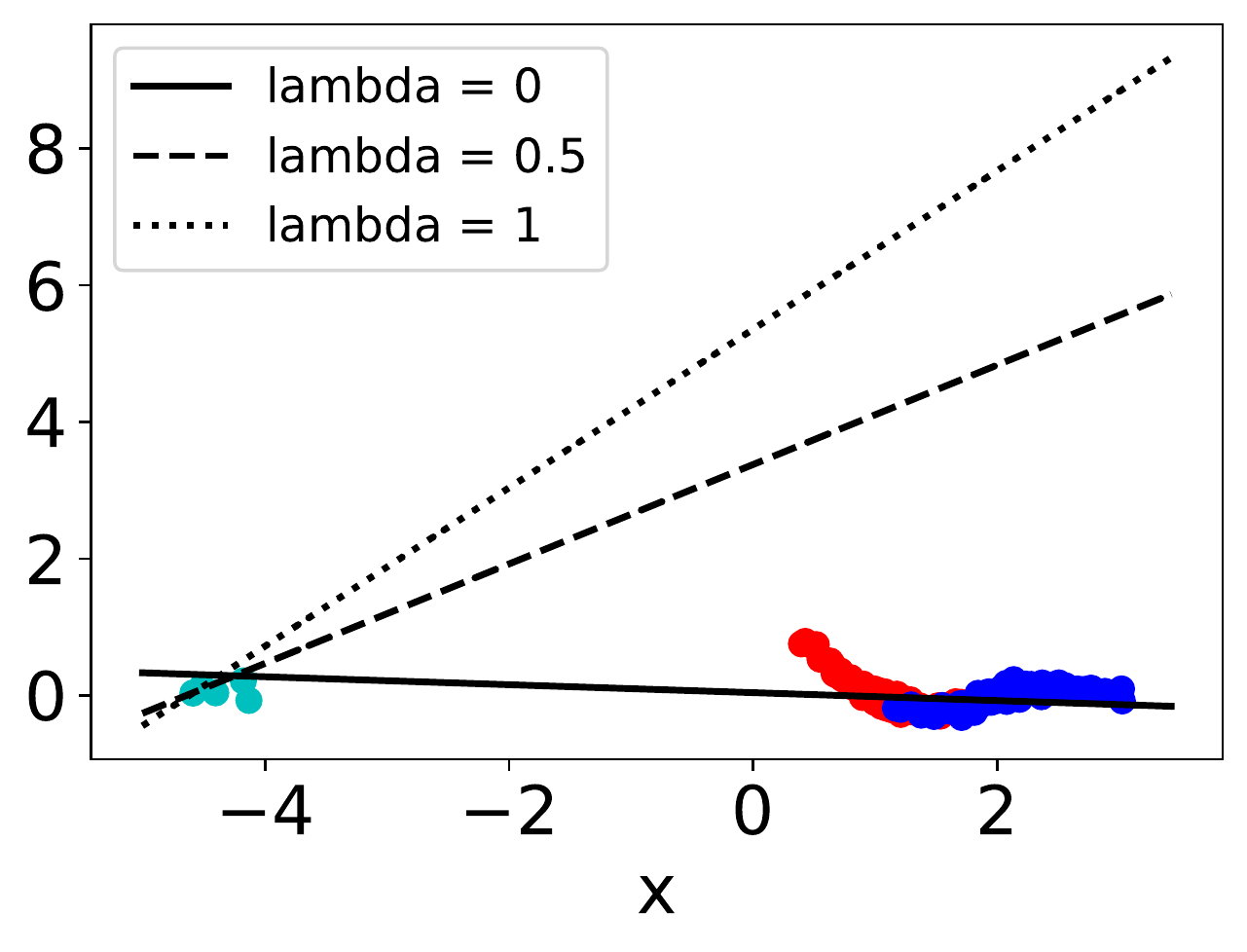}
    \caption{Gaussian}
    \label{noise_Gaussian}
  \end{subfigure}
  \begin{subfigure}{0.19\linewidth}
    \includegraphics[width = 1 \textwidth]{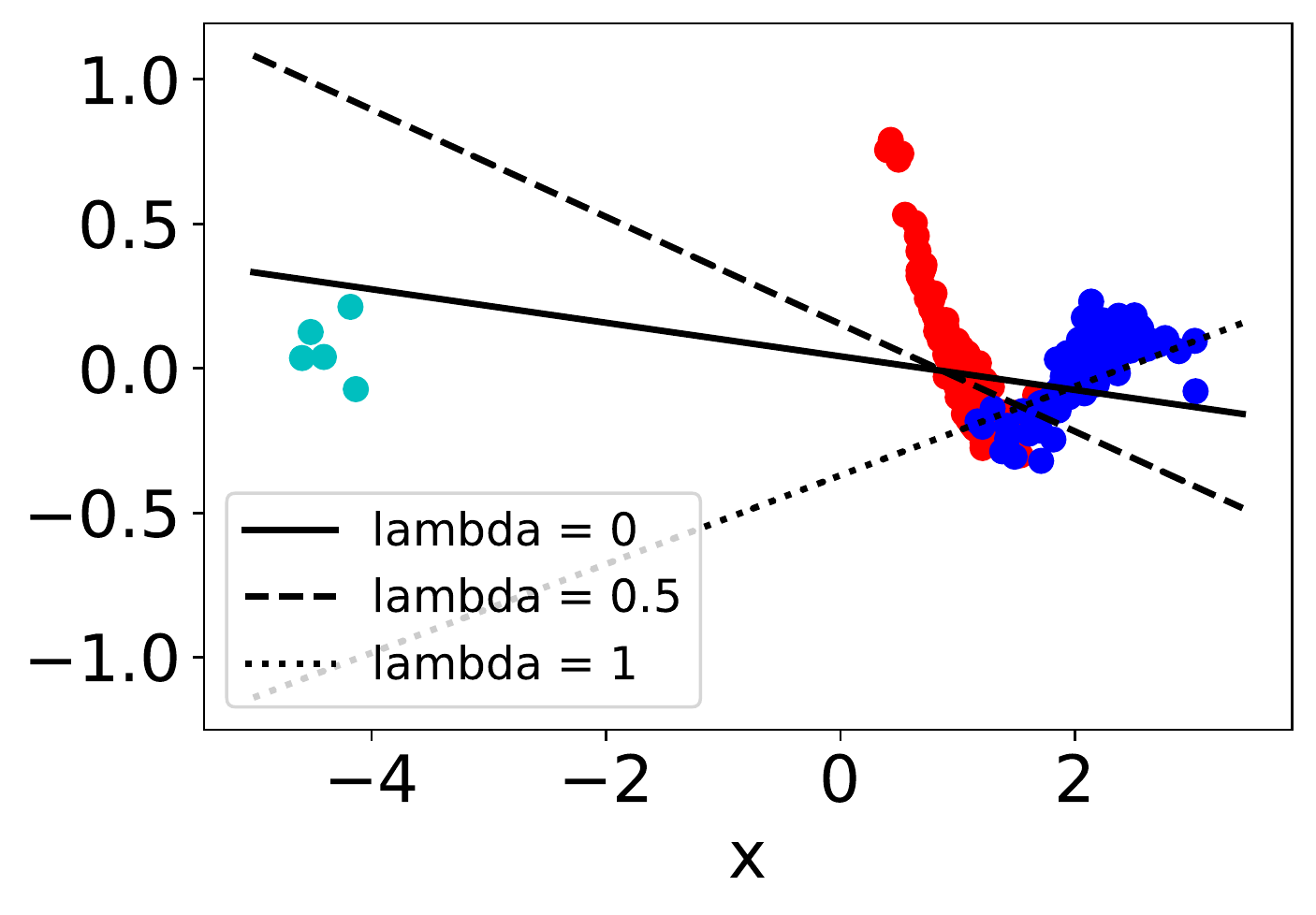}
    \caption{Student-t}
    \label{noise_t}
  \end{subfigure}
  \caption{(a) visualizes the marginal distribution of the source and target domain. The target function and the output value for the regression task are shown in (b). Figure (c) visualizes the fitted lines in the original scenario with Gaussian density estimation without outliers added. Figures (d) and (e) present the adaptation results of Gaussian density estimation and using Student's $t$-distribution with outliers separately. Lambda in the legend refers to the power $\lambda$ in Equation \eqref{eq: covariate weight}. Gaussian density estimation overfits to the outliers. }
\label{covariate_shift_regression}
\end{figure*}

Under covariate shift between a source domain $D_s$ and a target domain $D_t$, a fixed labelling function is assumed. We consider a standard weighting approach \cite{sugiyama2007covariate} to make the source domain distribution $p_{D_s}$ approximate the target domain distribution $p_{D_t}$:
\begin{equation}\label{eq: covariate weight}
    w_{cov} = \bigg(\frac{p_{D_t}(x)}{p_{D_s}(x)}\bigg)^\lambda.
\end{equation}
Here, $\lambda$ controls the strength of the weighting scheme. Similar to Section \ref{section: Bayes classifier}, Gaussian distributions $N(x|\mu; \sigma^2)$ with mean $\mu$ and variance $\sigma^2$ are estimated for $p_{D_s}$ and $p_{D_t}$. When outliers occur in the tail of the source distribution, extreme weights $w_{cov}$ are assigned to those outliers if the target distribution $p_{D_t}$ has a larger variance $\sigma^2_{D_t}$ than $\sigma^2_{D_s}$ of the source domain. This will lead to overfitting to these outliers only. With the use of a $t$-distribution, a weight of 1 is obtained, resulting in improved estimator behavior.

A domain adaptation regression setting is considered similar to the original work \cite{sugiyama2007covariate}. 
The target function is $f(x) = \text{sinc}(x)$, shown in Figure \ref{target_function}. The source and the target densities are
$p_{D_s}(x) = N(x| 1.1,(1/2)^2)$ and $p_{D_t}(x) = N(x| 2.1, (10/17)^2)$, respectively.
We add noise $\epsilon_{s_i}$ to the target function to create the output values for the source domain $y_{s_i} = f(x_{s_i}) + \epsilon_{s_i}$ with $p(\epsilon_{s})=N(\epsilon_{s}|0,(1/4)^2).$ We set the source sample size to $n_s=150$ and the target sample size to $n_t=100$. To approximate the target domain, each source sample is assigned a weight $w_{cov}$ according to Eq. \eqref{eq: covariate weight}. We randomly sample 5 outliers in the range $[-5, -4]$ and add them to the source domain after density estimation. All parameters are estimated by maximum likelihood. 
 
The sensitivities to outliers with Gaussian density estimation and using the $t$-distribution are compared in Figure \ref{covariate_shift_regression}. With Gaussian density estimation, noisy samples receive large weights due to the larger variance of the target domain, therefore the regressor overfits to the noisy samples when $\lambda$ is not 0. Student's $t$-distribution leads to small weights for the noisy samples because of its heavy tail. 
 
\subsubsection{Conservative Prediction}\label{section: conservative prediction}

We show that conservativeness leads to further desirable behavior under adversarial attacks. For networks trained with softmax, adversarial attacks make the network misclassify samples with high confidence. But when a model with softRmax is attacked, the sample is misclassified but with low confidence due to the conservativeness. 

We show the confidence of misclassified sample is low with softRmax by examining the posteriors of misclassified samples from the public dataset MNIST \cite{lecun1998mnist} under different levels of adversarial FGSM attacks. The network has four convolutional layers and one fully connected layer. We set a batchsize of $32$ and optimize the network by Adam with a learning rate of $1\text{e}-3$. We use the same architecture for the softmax and softRmax setting, with the only difference being the activation function after the final layer.

\begin{figure}[ht]
  \centering
  \begin{subfigure}{0.45\linewidth}
    \includegraphics[width = 1 \textwidth]{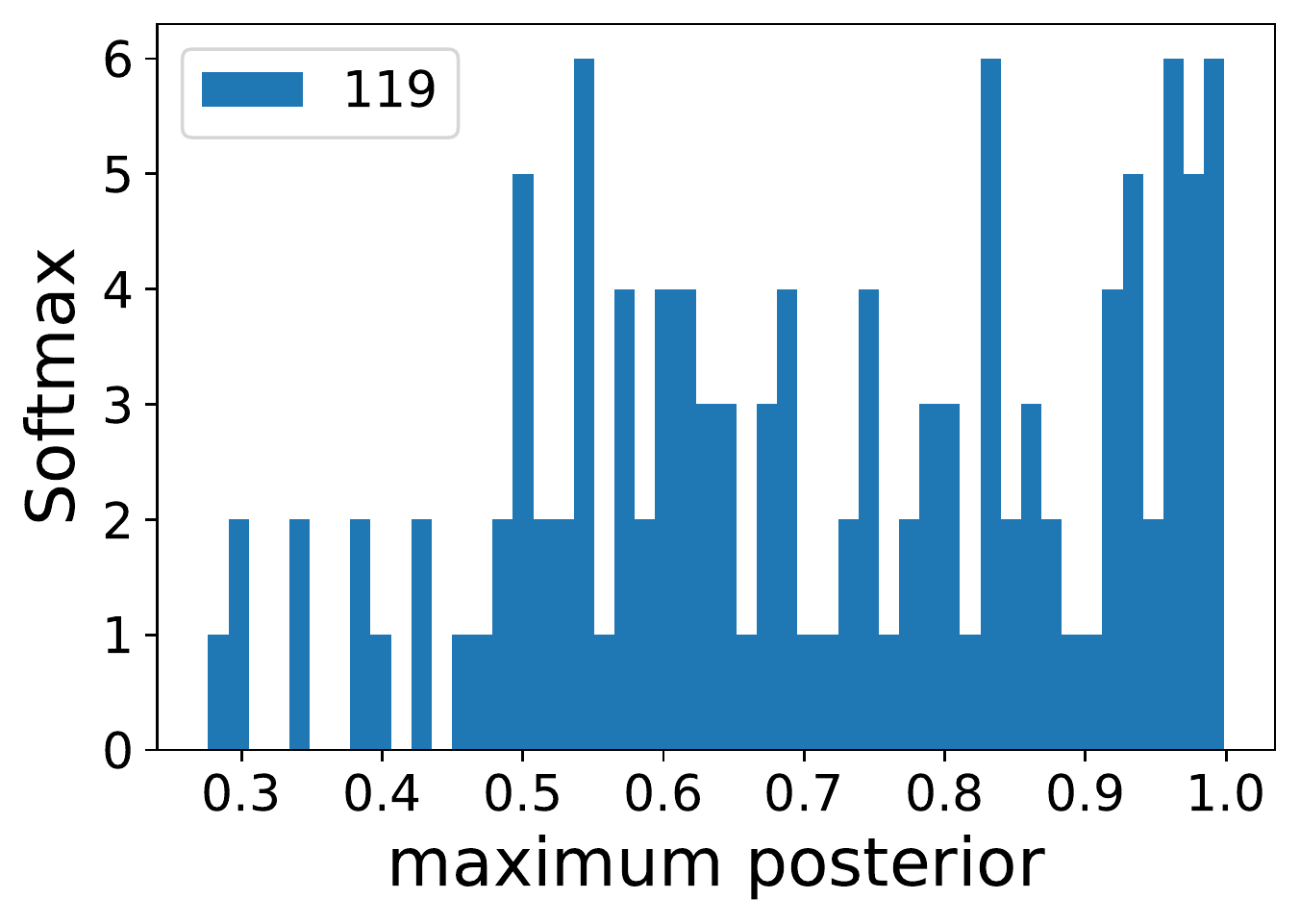}
    \caption{Softmax no attack}
     \label{subfig: pos_softmax_eps0}
  \end{subfigure}
  \begin{subfigure}{0.45\linewidth}
    \includegraphics[width = 1 \textwidth]{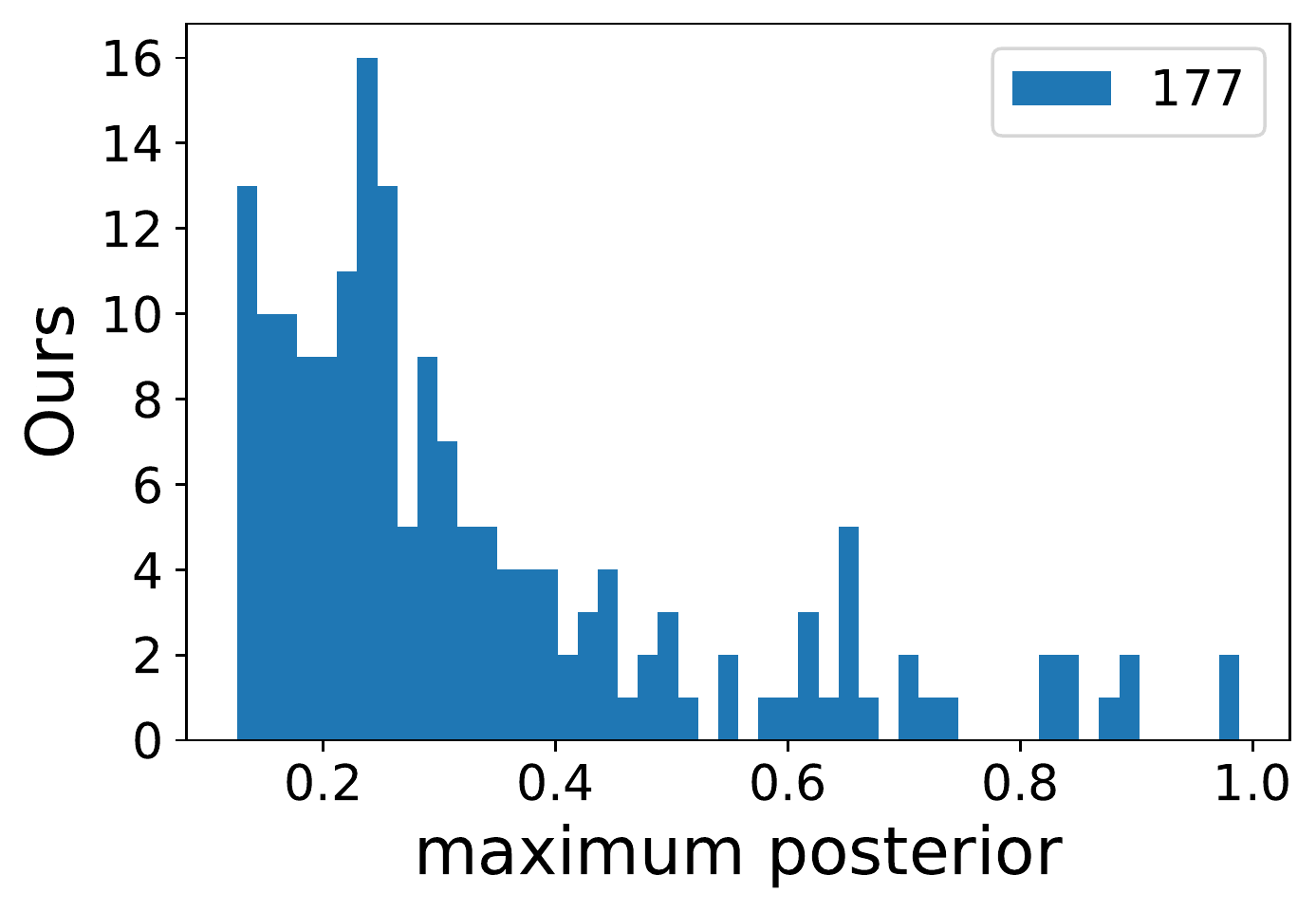}
    \caption{Ours no attack}
     \label{subfig: pos_marco_eps0}
  \end{subfigure}
  \begin{subfigure}{0.45\linewidth}
    \includegraphics[width = 1 \textwidth]{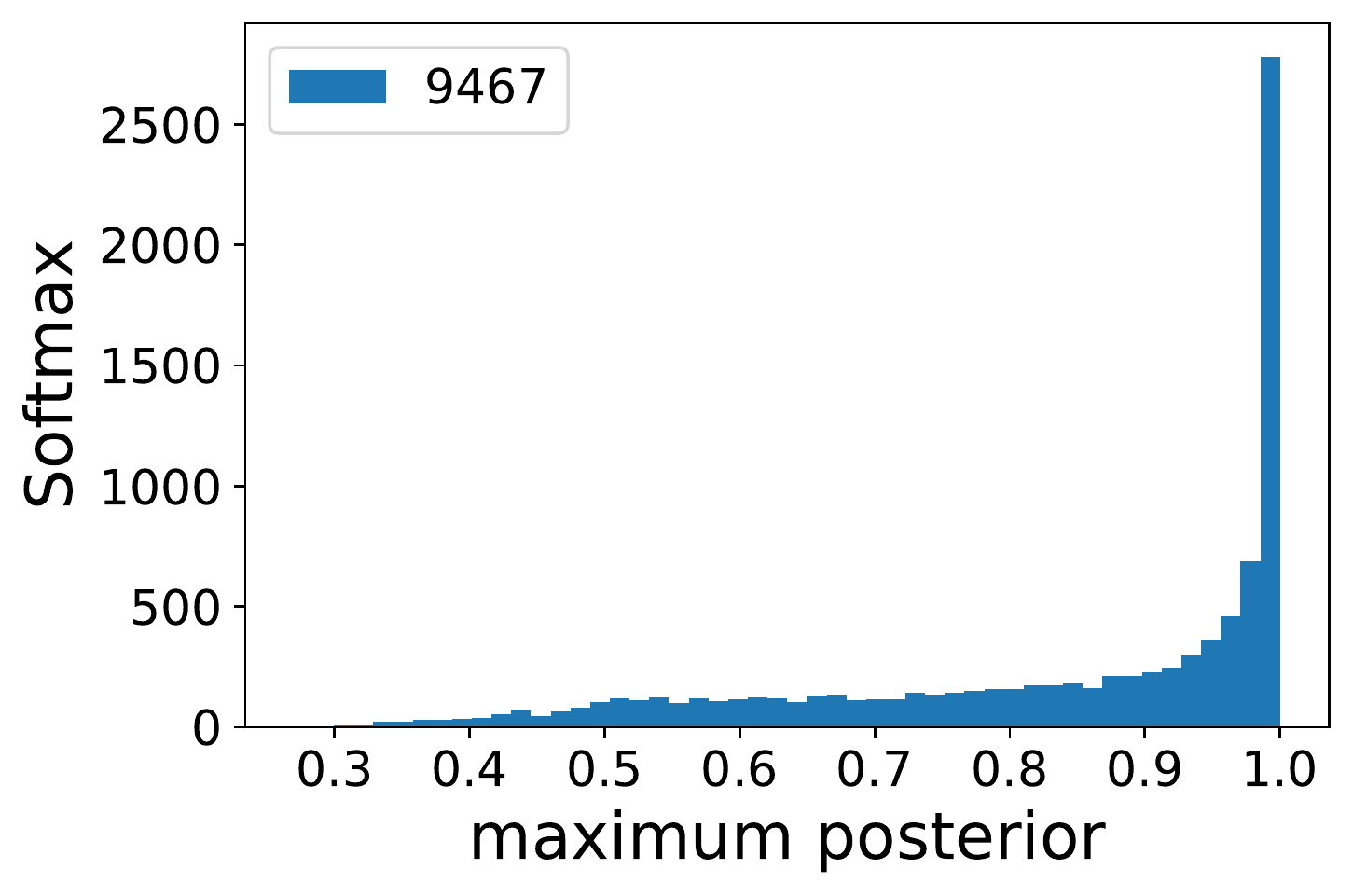}
    \caption{Softmax $\epsilon$=0.3}
     \label{subfig: pos_softmax_eps0.3}
  \end{subfigure}
  \begin{subfigure}{0.45\linewidth}
    \includegraphics[width = 1 \textwidth]{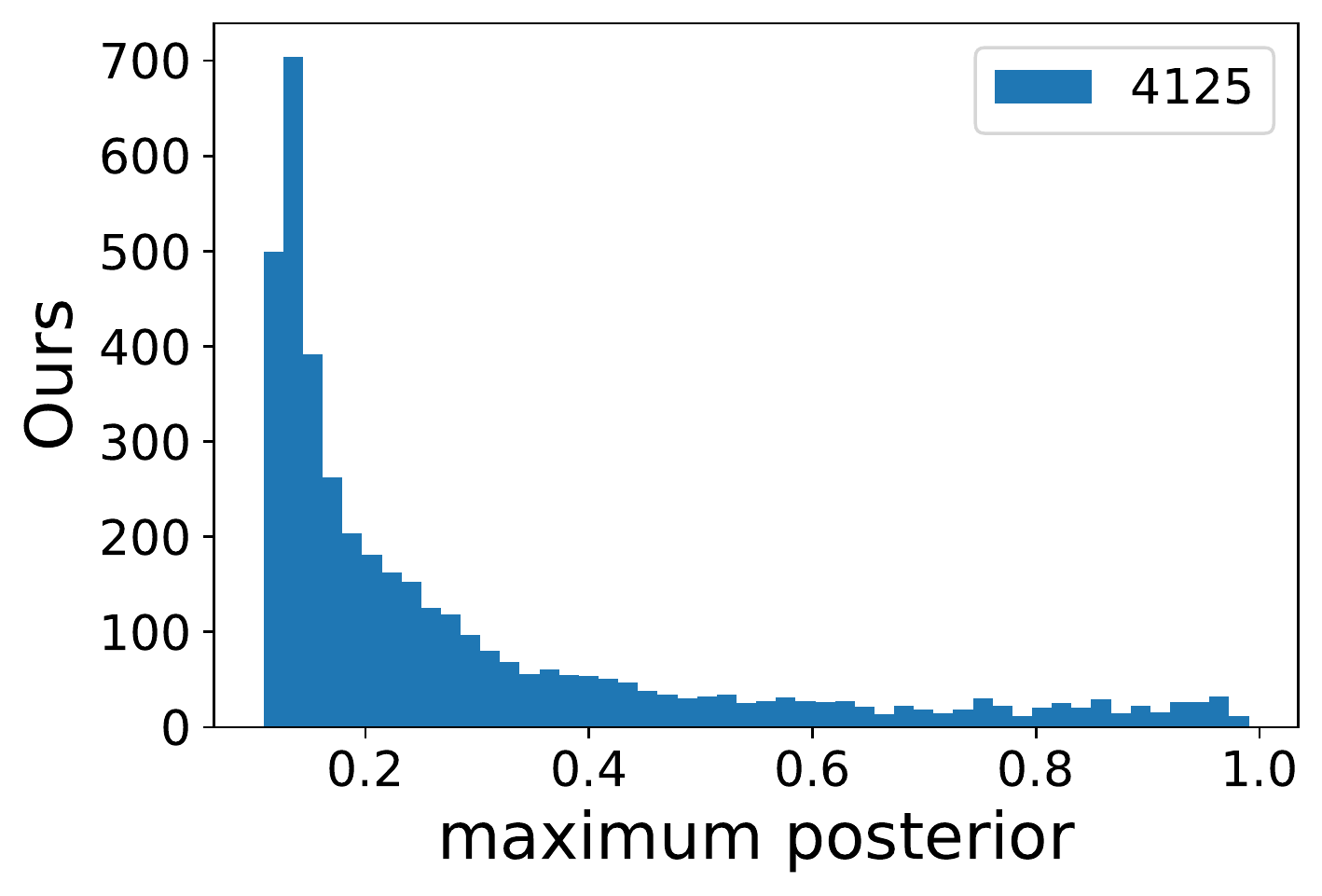}
    \caption{Ours  $\epsilon$=0.3}
     \label{subfig: pos_marco_eps0.3}
  \end{subfigure}
  \begin{subfigure}{0.45\linewidth}
    \includegraphics[width = 1 \textwidth]{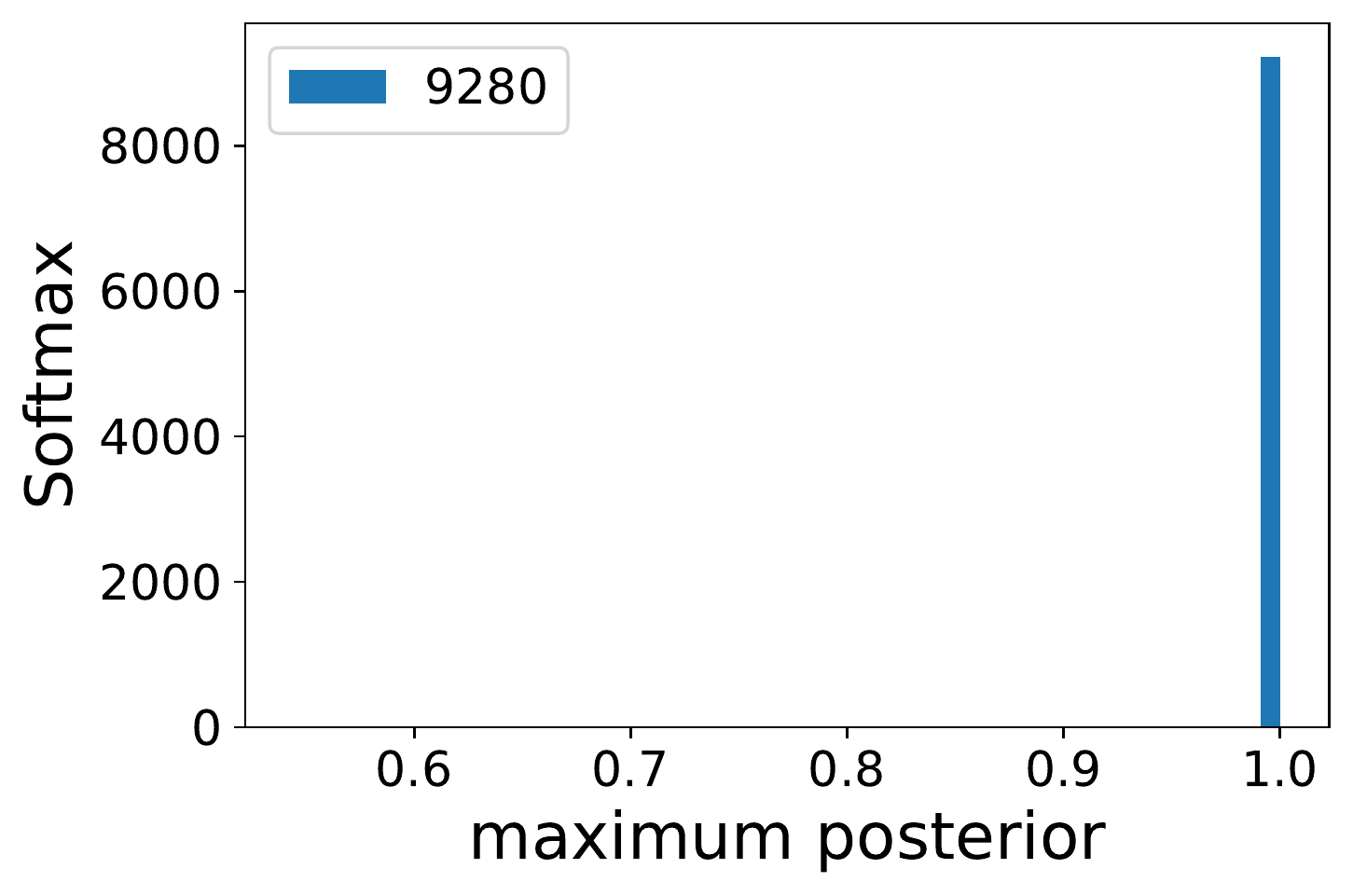}
    \caption{Softmax $\epsilon$=100}
     \label{subfig: pos_softmax_eps100}
  \end{subfigure}
   \begin{subfigure}{0.45\linewidth}
    \includegraphics[width = 1 \textwidth]{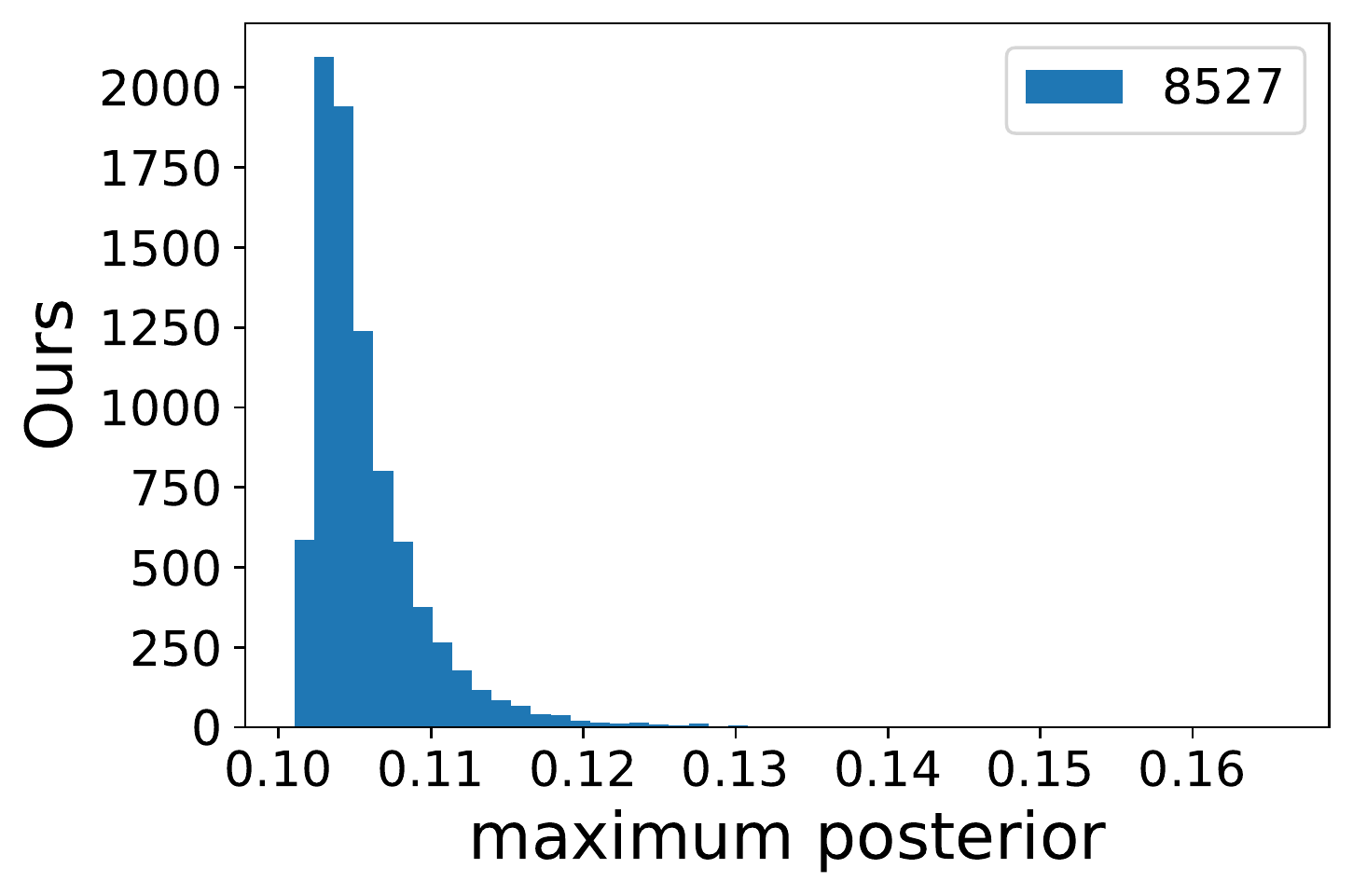}
    \caption{Ours  $\epsilon$=100}
     \label{subfig: pos_marco_eps100}
  \end{subfigure}
  \caption{Posteriors of predicted class for misclassified MNIST test samples under different levels of FGSM attacks. The legends give the number of misclassified samples. Misclassified samples in the setting of softRmax receive less confident prediction. Even under extreme attacks with $\epsilon=100$, our softRmax gives non-saturated posteriors at random-guess level.}
\label{fig: posteriors}
\end{figure}

As shown in Figure \ref{fig: posteriors}, for the network trained with the standard softmax, posteriors on the predicted class of misclassified samples are high on average. Specifically, using softmax, under large scale attacks with $\epsilon=100$, all samples are misclassified with a posterior of 1. Due to the conservativeness in the tail and the soft posterior change, our softRmax leads to posteriors around random-guess level. 

\subsection{Robustness}

\subsubsection{Adversarial Defense}\label{subsubsection: public datasets}

We perform experiments on public datasets MNIST \cite{lecun1998mnist}, CIFAR10 \cite{Krizhevsky09learningmultiple}, and CIFAR100 with standard softmax and our softRmax. 
The setting of MNIST is the same as in Section \ref{section: conservative prediction}. A randomly initialized VGG16 network is used for CIFAR10 classification. We optimize VGG16 by SGD with a learning rate of $5\text{e}-3$, batchsize $256$ and weight decay $5\text{e}-6$. For CIFAR100, we adopt ResNet50 pretrained on ImageNet and finetune it with Adam. We set the learning rate to be $1\text{e}-4$, batchsize $512$ and weight decay $5\text{e}-6$. Note that no extra data augmentation is used in any experiment. The only difference between the baseline with softmax and our approach is the activation function after the final fully connected layer. By simply substituting the softmax activation function with the polynomial softRmax activation function, the network develops strong adversarial defense ability (see Table \ref{tab: adversarial attack}). Without being combined with other approaches, the naive softRmax model can outperform state of the art adversarial defense approaches based on attention mechanism on CIFAR datasets \cite{agrawal2021impact}.
\begin{table}[h]
    \caption{Adversarial defense results. We compare networks with softmax and our softRmax activation under FGSM and BIM attacks ($T$=10). `Clean' refers to the classification accuracy on the testset without any attack. We consider binary classification for class 3 and 7 from MNIST, MNIST, CIFAR10, and CIFAR100 under different attack levels $\epsilon$. The results show a clear improvement of the robustness to adversarial attacks with softRmax.}
    \label{tab: adversarial attack}
    \centering
    \resizebox{0.47\textwidth}{!}{\begin{tabular}{lllllll}
        \toprule
        \multirow{2}{*}{Dataset} & \multirow{2}{*}{Method} &\multirow{2}{*}{Clean}
       &  \multicolumn{2}{c}{FGSM} &   \multicolumn{2}{c}{BIM} \\
        \cmidrule{4-7}
        {} & {} &{} &$\epsilon$=0.1 &$\epsilon$=0.3 &$\epsilon$=0.1 &$\epsilon$=0.3 \\
        \midrule
        \multirow{2}{*}{MNIST 3$\&$7} &softmax &99.75 &77.18&18.69  &71.64& 0.24 \\
        {}&ours &\textbf{99.95} &\textbf{95.88}& \textbf{88.71}  &\textbf{94.90}& \textbf{66.24}\\
        \cmidrule{2-7}
        \multirow{2}{*}{MNIST} &softmax &96.8 &48.3&0.41  &53.49&0.02 \\
        {}&ours &\textbf{97.78} &\textbf{75.55}&\textbf{49.30} &\textbf{69.73}&\textbf{33.94}\\
        \cmidrule{2-7}
        \multirow{2}{*}{CIFAR10} &softmax &\textbf{80.31} &17.62&13.95 &10.39&4.83 \\
        {}&ours &80.28 & \textbf{49.93}&\textbf{41.03} & \textbf{44.25}&\textbf{18.11}\\
        \cmidrule{2-7}
        \multirow{2}{*}{CIFAR100} &softmax &\textbf{61.43} &11.23&6.78 &1.94&0.05 \\
        {}&ours &61.04 & \textbf{19.31}&\textbf{11.04} & \textbf{9.06}&\textbf{2.16}\\
        \bottomrule
    \end{tabular}}
\end{table}



\subsubsection{Gradient Obfuscation}\label{section: exp_gradient_obfuscation}
As we noted in Section \ref{section: robustness from the tail}, different from gradient obfuscation, our loss landscape is not rough but simply has a different structure in the tail of the distribution. Leading a sample to the tail is different from blocking the attack by local minimum due to a rough loss landscape. The tail is the right direction to perturb the sample from the point of view of the gradient based attacks because the overall loss monotonically increases towards the tail.

\begin{figure}[h]
  \centering
  \begin{subfigure}{0.49\linewidth}
    \includegraphics[width = 1 \textwidth]{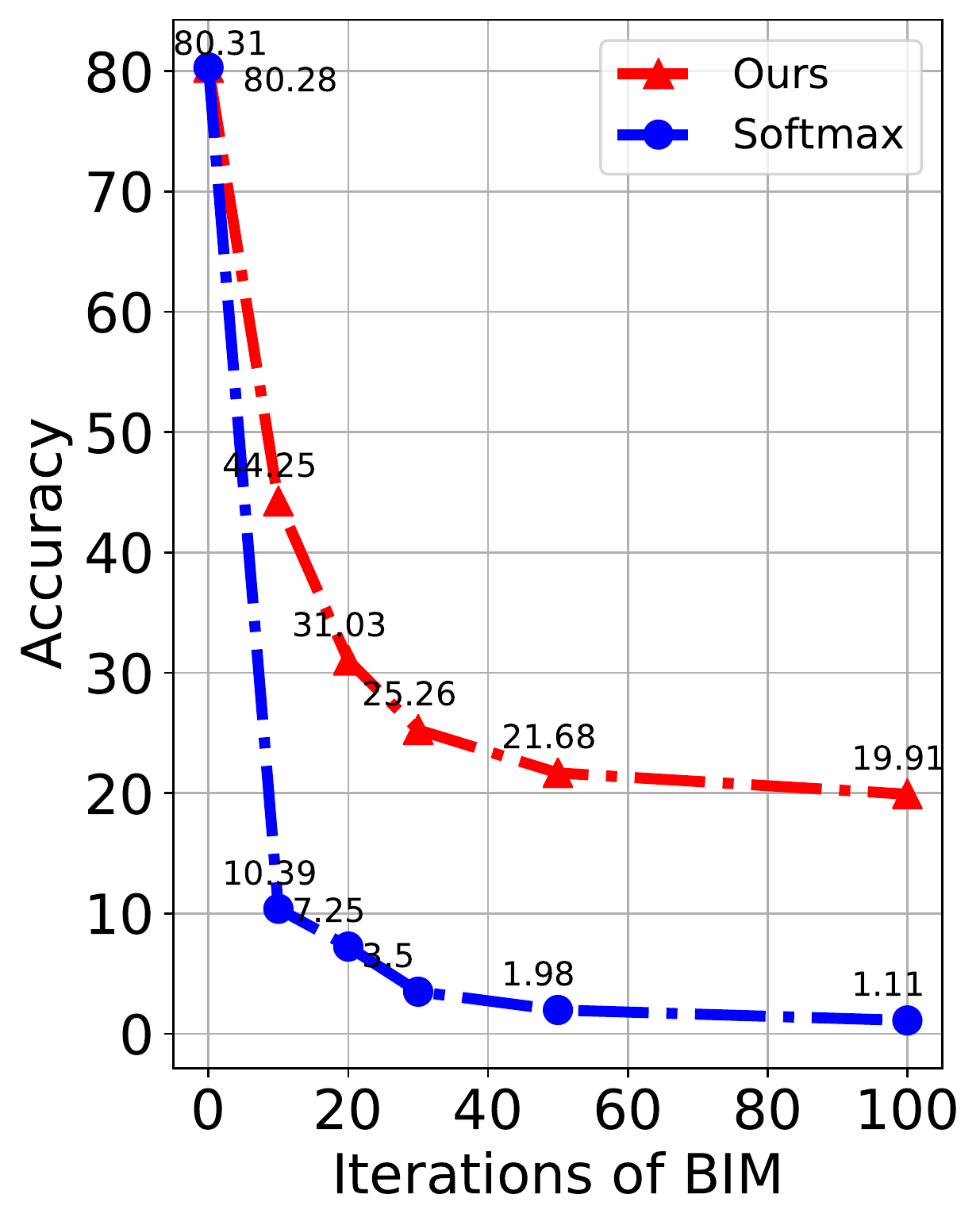}
    \caption{$\epsilon$=0.1}
     \label{subfigure: BIM_0.1}
  \end{subfigure}
  \begin{subfigure}{0.49\linewidth}
    \includegraphics[width = 1 \textwidth]{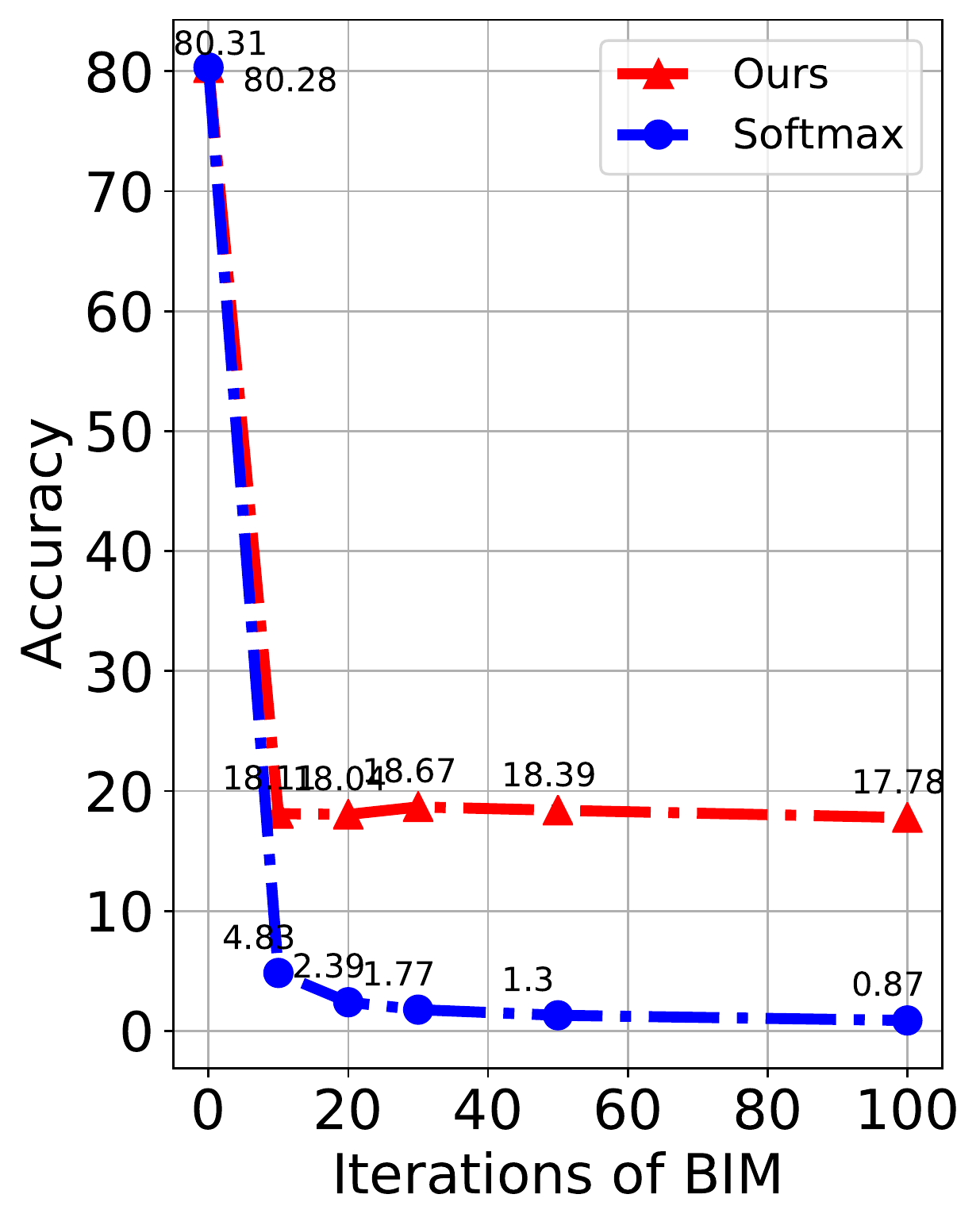}
    \caption{$\epsilon$=0.3}
    \label{subfigure: BIM_0.3}
  \end{subfigure}
  \caption{BIM attack on CIFAR10 with different iteration numbers. For both attack levels $\epsilon=0.1$ and $\epsilon=0.3$, the stabilized accuracy of softRmax is significantly higher than that of softmax.}
\label{figure: BIM iterations}
\end{figure}

Nevertheless, we also rule out the possibility of gradient obfuscation by performing the iterative BIM attack at very large iteration numbers. In fact, softRmax shows stronger robustness after the accuracy stabilises with increased iterations, as shown in Figure \ref{figure: BIM iterations}. The experiment in the next section shows that the black-box attack is a weaker attack than the white-box attack, which further diminishes the possibility that the softRmax robustness can be explained by gradient obfuscation.
\subsubsection{Robustness from Conservativeness}
Existing gradient-based attacks can only examine the robustness of a model as a whole but cannot show whether the robustness comes from the enlarged margin or from the conservativeness or both. To check the effect of the enlarged margin and the conservative tail on robustness, we propose a semi-black-box attack, coined the average-sample attack. It does not rely on the gradient but simply perturbs a sample to the direction of a selected target class based on a precomputed average, so the sample is guaranteed to not be pushed towards the tail. We precompute the average sample $\mathbf{Avg}_y= \frac{1}{n}\sum^n \mathbf{x}_{ny}$ for each class $y$. With $t$ the target class, the adversarial input $\mathbf{x}'$ then becomes
\begin{equation}
\mathbf{x}'_y = \mathbf{x}_y + \epsilon\text{sign}(\mathbf{Avg}_t - \mathbf{Avg}_y).
\end{equation}

In general, our average-sample attack should not be stronger than the gradient-based ones because the attacking direction found by the former attack is not optimized. It prevents the sample from going to the tail, so if it becomes a stronger attack for the softRmax, it indicates that the conservativeness in the tail indeed gives added robustness. Otherwise the gradient based white-box attack should be the worst attack for our softRmax model as well.

\begin{table}[hbt]
    \caption{Results of targeted attack on MNIST dataset. White refers to the targeted attack with the network available for generating adversarial samples. Black is the black-box version of the targeted attack, where a substitute network is first learned to approximate the decision boundary of the original model. Avg is our average-sample attack. The column Clean is the original per class accuracy of different models without any adversarial attack. The rest results are the accuracy of all the 10 classes under adversarial attacks. We highlight the \textbf{worst} performance of each model among all attacks.} 
    \label{tab: targeted attack}
    \centering
    \resizebox{0.47\textwidth}{!}{\begin{tabular}{lllllllll}
        \toprule
        \multirow{2}{*}{Classes} & \multicolumn{2}{c}{Clean}
       & \multicolumn{2}{c}{White} &   \multicolumn{2}{c}{Black} & \multicolumn{2}{c}{Avg}\\
        \cmidrule{2-9}
        {} &softmax &Ours &softmax &Ours &softmax &Ours &softmax &Ours \\
        \midrule
        0 &98.88 &99.18 &\textbf{11.14}  &\textbf{53.2} &33.55 &74.3 &30.69 &58.13\\
        \cmidrule{2-9}
        1 &98.50 &99.21 &\textbf{15.05} &\textbf{46.73} &53.41 &60.73 &48.77 &63.44\\
        \cmidrule{2-9}
        2 &98.26 &98.16 &\textbf{10.50} &54.00 &25.93 &50.42 &16.53 &\textbf{42.26}\\
        \cmidrule{2-9}
        3 &96.34 &98.12 &\textbf{10.31} &51.73 &27.16 &58.99 &15.54 &\textbf{37.77}\\
        \cmidrule{2-9}
        4 &96.84 &97.35 &\textbf{13.28} &51.84 &37.21 &63.03 &26.97 &\textbf{47.94}\\
        \cmidrule{2-9}
        5 &97.87 &98.09 &\textbf{9.33} &52.24 &32.84 &64.52 &16.68 &\textbf{46.76}\\
        \cmidrule{2-9}
        6 &97.91 &98.64 &\textbf{12.00} &51.52 &35.46 &52.14 &24.43 &\textbf{44.56}\\
        \cmidrule{2-9}
        7 &96.60 &96.69 &\textbf{12.11} &52.88 &32.46 &60.10 &22.03 &\textbf{45.50}\\
        \cmidrule{2-9}
        8 &92.61 &96.61 &\textbf{9.36} &54.38 &24.96 &73.04 &18.90 &\textbf{41.93}\\
        \cmidrule{2-9}
        9 &94.05 &95.64 &\textbf{6.33} &51.72 &30.65 &68.58 &29.31 &\textbf{49.67}\\
 
        \bottomrule
    \end{tabular}}
\end{table}

To check whether the average-sample attack is a stronger attack on the softRmax model, we also perform a gradient based targeted attack in both the white-box setting and the black-box setting. The latter one checks whether gradient obfuscation happens. In the black-box attack \cite{papernot2017practical}, a substitute model that is used to generate adversarial samples is first learned to mimic the decision boundary of the original model. We show that the white-box attack is the strongest attack for softmax while our customized average-sample attack is more effective on the softRmax (see Table \ref{tab: targeted attack}). This indicates that the conservative tail of softRmax indeed leads to robustness. The fact that the black-box attack is weaker than the white-box attack eliminates the possibility that the weaker performance of average-sample attack is brought by gradient obfuscation. Also, even when the average-sample attack gives the lowest accuracy on softRmax, its performance is still significantly better than that of softmax.

\subsubsection{Robustness from Enlarged Margin}

We further demonstrate that the robustness of softRmax also comes from the enlarged margin. If all samples are pushed to the decision boundary instead of the tail, then the model with a larger margin is more robust. It is hard to measure the margin in the input space 
so we use the prediction margin $M_z$, as specified in Equation \eqref{eq: prediction margin}, as alternative. If the perturbation in $\mathbf{x}$ can push the sample across the margin, then it means the corresponding change in $M_z$ is also larger than the original prediction margin $M_z$. 
However, $M_z$ cannot be used to measure the margin directly due to different mappings from $\mathbf{x}$ to $\mathbf{z}$ of different models.  A larger $M_z$ does not imply a larger margin in the input space. So we introduce a new metric, the magnitude-margin ratio, to measure the change in $M_z$ caused by an attack with respect to the original prediction margin. If the change is larger than the original prediction margin for a sample, it indicates that this sample can be successfully attacked.

\begin{figure}[hbt]
  \centering
     \begin{subfigure}{0.45\linewidth}
    \includegraphics[width = 1 \textwidth]{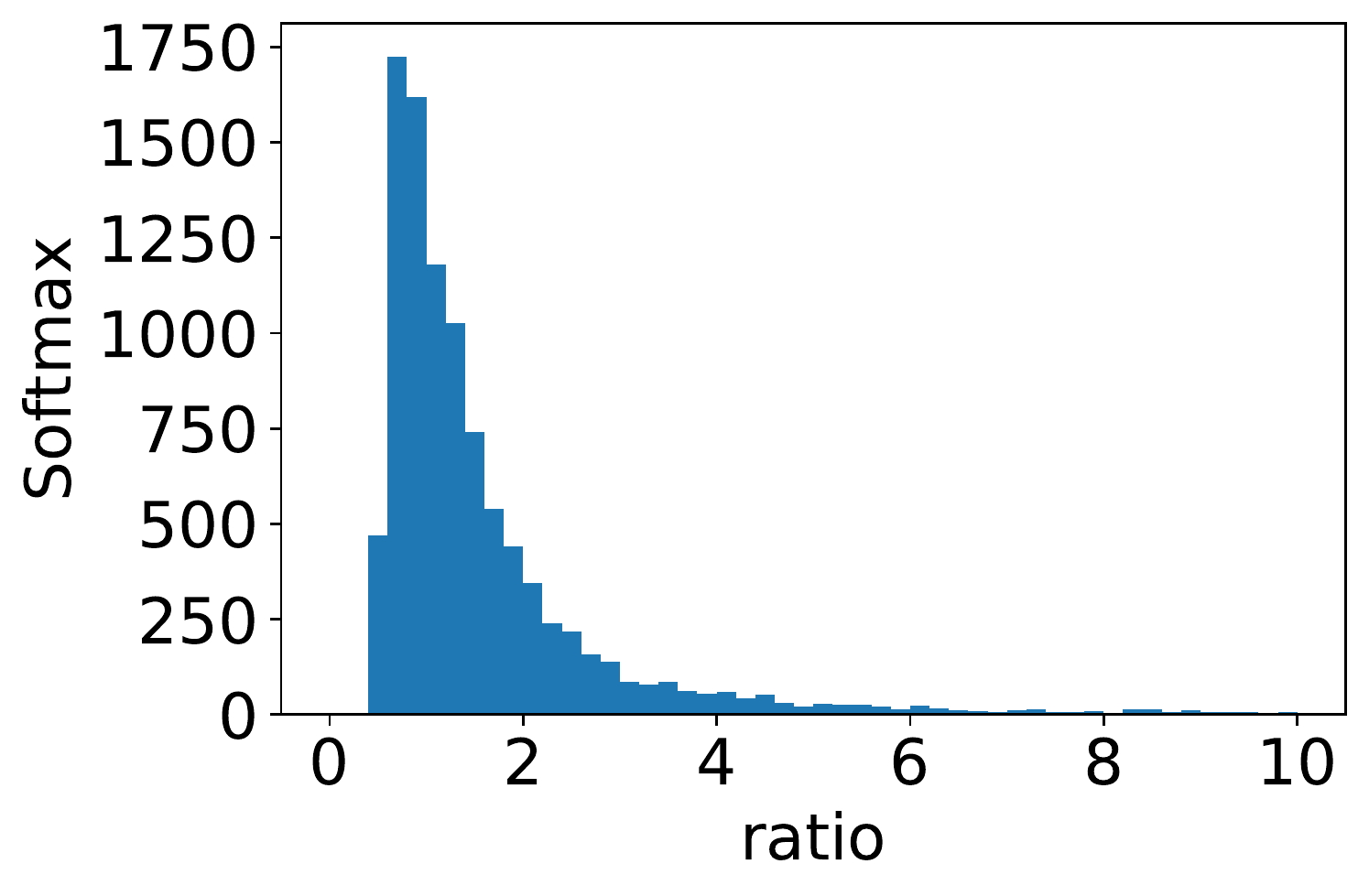}
    \caption{MNIST}
     \label{subfig: softmax_mnist}
    \end{subfigure}
    \begin{subfigure}{0.45\linewidth}
    \includegraphics[width = 1 \textwidth]{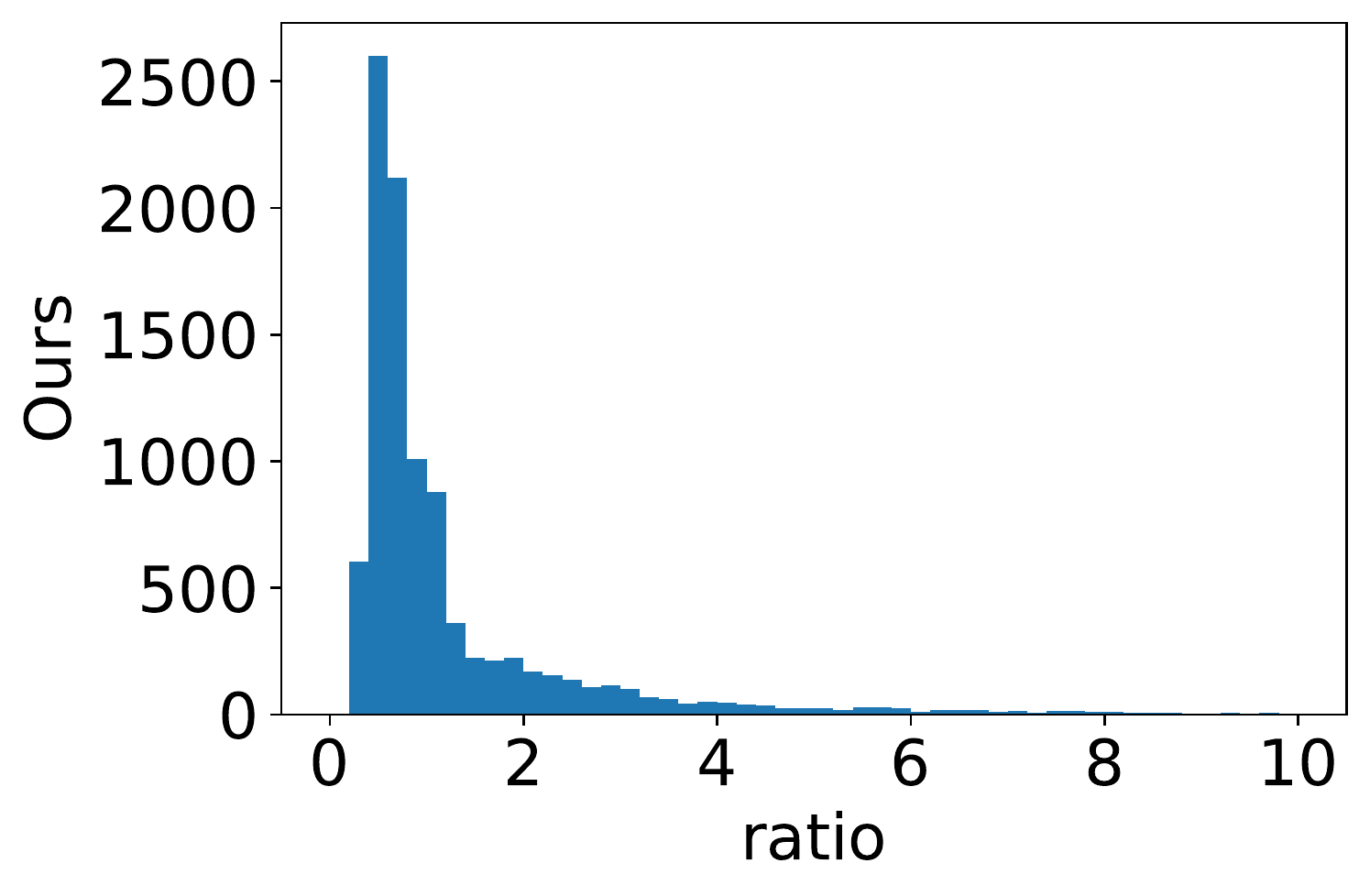}
    \caption{MNIST}
     \label{subfig: marco_mnist}
  \end{subfigure}
  \begin{subfigure}{0.45\linewidth}
    \includegraphics[width = 1 \textwidth]{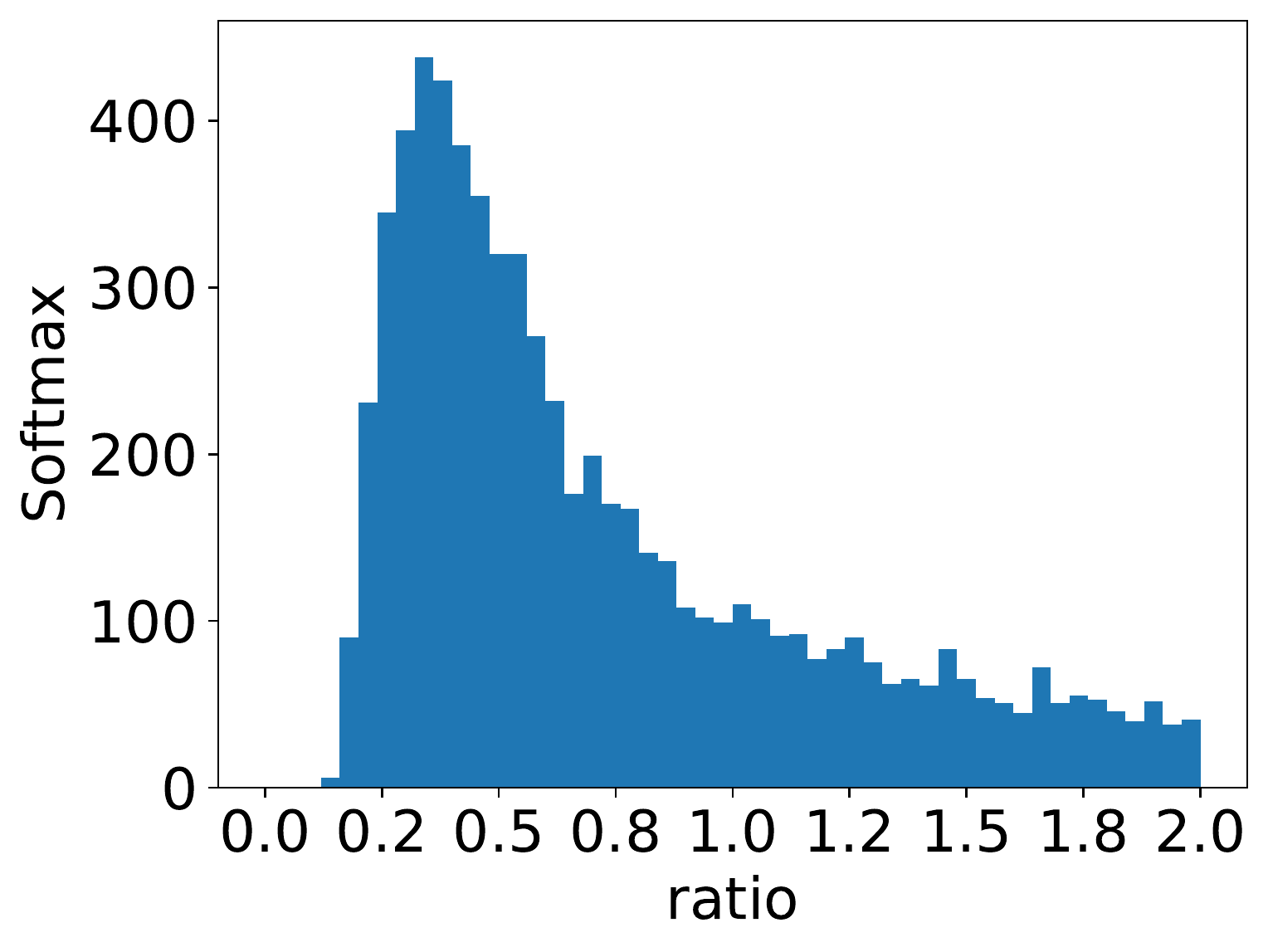}
    \caption{CIFAR10}
     \label{subfig: softmax_cifar10}
  \end{subfigure}
  \begin{subfigure}{0.45\linewidth}
    \includegraphics[width = 1 \textwidth]{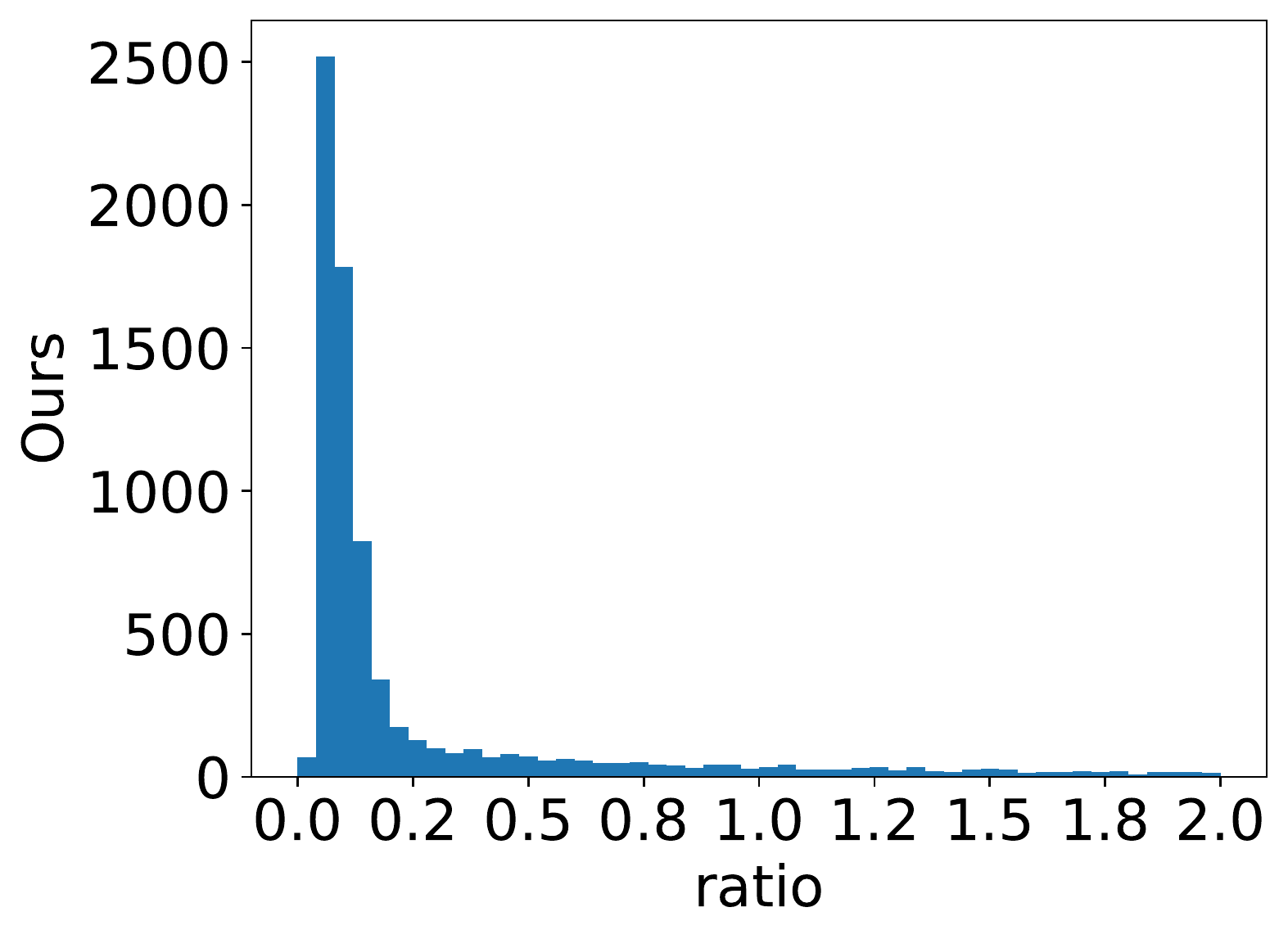}
    \caption{CIFAR10}
     \label{subfig: marco_cifar10}
    \end{subfigure} 
    \begin{subfigure}{0.45\linewidth}
    \includegraphics[width = 1 \textwidth]{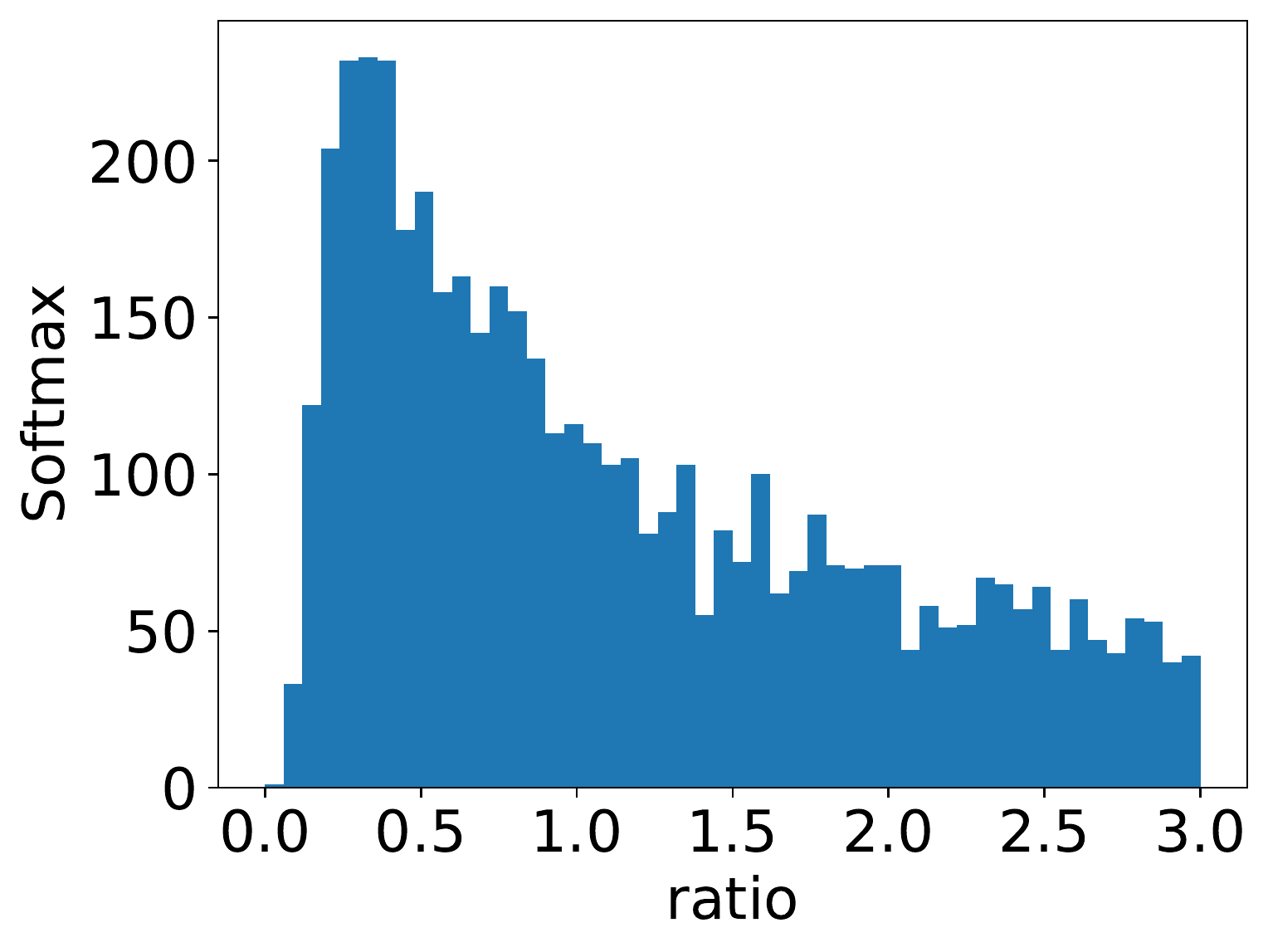}
    \caption{CIFAR100}
     \label{subfig: softmax_cifar100}
  \end{subfigure}
  \begin{subfigure}{0.45\linewidth}
    \includegraphics[width = 1 \textwidth]{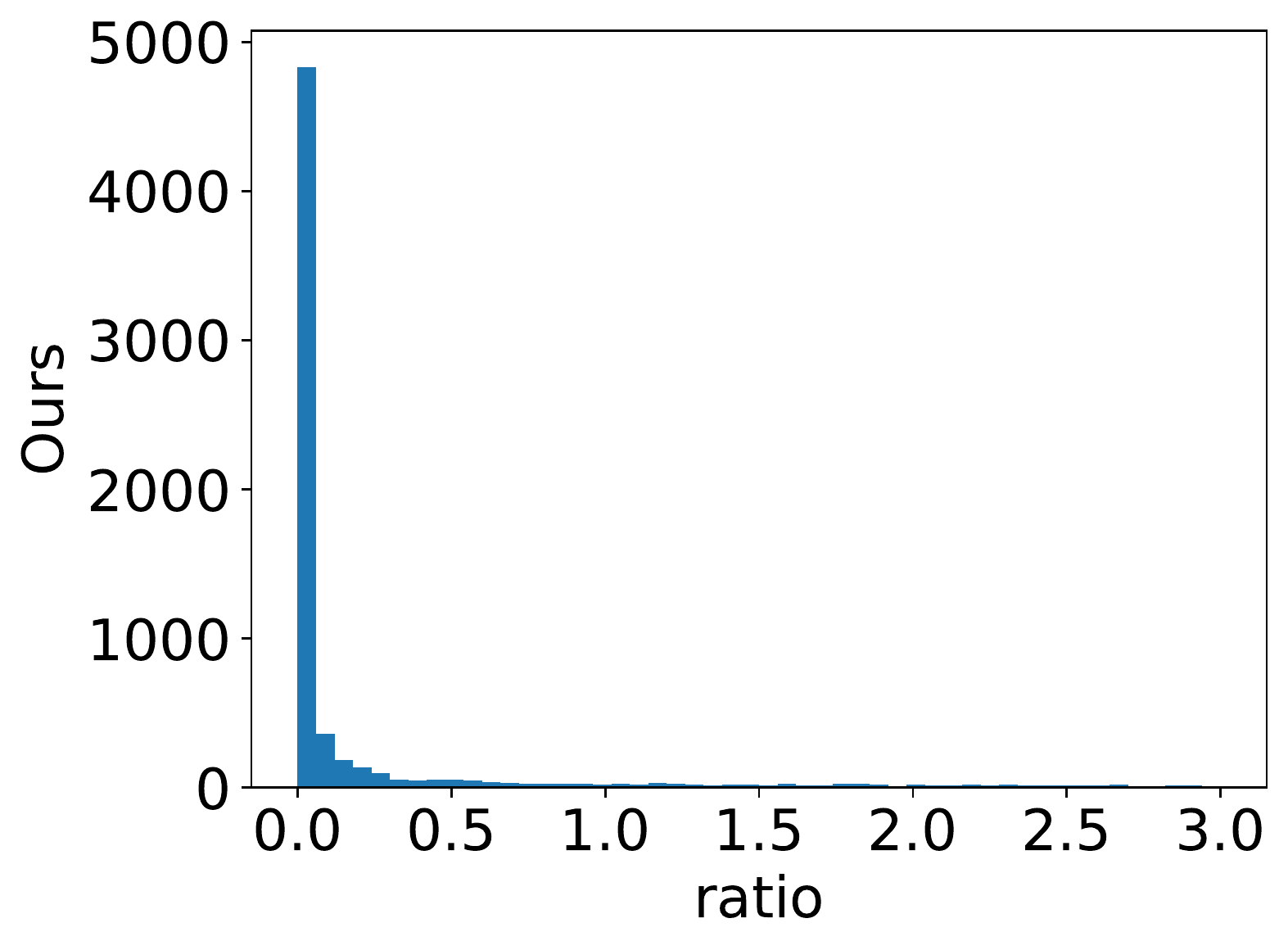}
    \caption{CIFAR100}
     \label{subfig: marco_cifar100}
    \end{subfigure} 
  \caption{Histogram of the magnitude-margin ratio $R$ of softmax and softRmax on all test data from MNIST, CIFAR10, and CIFAR100. The ratio of softRmax is significantly smaller, which indicates a higher robustness against adversarial attacks.}
\label{fig: ratio}
\end{figure}

To derive the ratio for an $\mathbf{x}$, we assume the index for $\text{max}_{i\neq y}\{\mathbf{z}_i\}$ is $j$. We denote the gradient of $\mathbf{z}_y$ and $\mathbf{z}_j$ of the input $\mathbf{x}$ by $\mathbf{w}_y$ and $\mathbf{w}_j$, respectively. After adding a perturbation $\mathbf{\eta}$ in the input $\mathbf{x}$, $\mathbf{z}_y$ and $\mathbf{z}_j$ change to $\Tilde{\mathbf{z}_y}$ and $ \Tilde{\mathbf{z}_j}$, where $ \Tilde{\mathbf{z}_y} = \mathbf{w}_y^T\mathbf{x} + \mathbf{w}_y^T\mathbf{\eta}$. The new prediction margin $\Tilde{M_z} = \Tilde{\mathbf{z}_y} - \Tilde{\mathbf{z}_j}$. According to \cite{goodfellow2015explaining}, $\mathbf{w}^T\mathbf{\eta}$ can be approximated by the magnitude $m$ of gradients, the attacking level $\epsilon$ of $\mathbf{\eta}$, and the dimension $n$ of input as $\epsilon mn$ and so
\begin{equation}
   r =\frac{|\Tilde{M_z} - M_z|}{M_z}
    = \frac{|(\mathbf{w}_y^T - \mathbf{w}_j^T)\mathbf{\eta}|}{M_z}\approx \frac{\epsilon mn}{M_z}.
\end{equation}
Given the same input dimension $n$ and attacking level $\epsilon$, the simplified ratio $R = \tfrac{m}{M_z}$ 
can serve as the metric. A model with a distribution of lower ratio $R$ means that, with the same level of attack, it is harder to change the prediction margin $M_z$, which indicates a larger margin in the input space and a higher robustness of this model. Figure \ref{fig: ratio} shows that the model with softRmax has ratios $R$ lower than softmax has on MNIST, CIFAR10, and CIFAR100.

%% file: conclusion.tex
\section{Discussion and Conclusion}



We suggest an easy substitution of polynomiality for exponentiality in several scenarios, showing it leads to conservative behavior regarding samples in the tail of the distribution.
For our polynomial softRmax, this behavior also leads to increased robustness against adversarial attacks.  We show that the robustness of softRmax also comes from an enlarged margin and link this to the inherent gradient regularization of softRmax, which demonstrates that its success does not stem from gradient obfuscation.  Our softRmax can be combined readily with many other adversarial defense strategies and it would be of interest to study their combined strength. 

Given the type of conservative behavior polynomiality induces, it seems worthwhile to study its usage in OOD detection and other problems related to non-i.i.d. sampling, domain adaptation, etc.

As for softmax, good DNN weight initialization is important to avoid gradient vanishing for softRmax. Apart from that, considering the current level of understanding and the experimental evidence provided, we see no restrictions to its usage.  In conclusion: why not give softRmax a try?





%% file: acknowledgement.tex
\paragraph{Acknowledgements.}
Many thanks to Jan van Gemert for his feedback and invaluable help with the rebuttal. Funded in part by the Netherlands Organization for Scientific Research (NWO), research program C2D–Horizontal Data Science for Evolving Content (628.011.002).